\documentclass[twoside,11pt]{article}

\usepackage[abbrvbib, preprint]{jmlr2e}
\usepackage{amsmath}
\usepackage{subfigure}
\usepackage{algorithmic,algorithm}
\usepackage{amsfonts}
\usepackage{bm}
\usepackage{hyperref} 
\newtheorem{assum}{Assumption} 
\usepackage{etoolbox}
\usepackage{color}
\makeatletter
\patchcmd{\@makecaption}
{\scshape}
{}
{}
{}
\makeatother

\usepackage{lastpage}
\jmlrheading{22}{2021}{1-\pageref{LastPage}}{1/20}{5/21}{20-070}{Ping Xu and Yue Wang and Xiang Chen and Zhi Tian}
\ShortHeadings{COKE: Communication-Censored Decentralized Kernel Learning}{Xu and Wang and Chen and Tian}
\firstpageno{1}

\begin{document}

\title{COKE: Communication-Censored Decentralized Kernel Learning}

\author{\name Ping Xu \email pxu3@gmu.edu 
       \AND
       \name Yue Wang \email ywang56@gmu.edu 
       \AND
       \name Xiang Chen \email xchen26@gmu.edu
       \AND
       \name Zhi Tian \email ztian1@gmu.edu \\
       \addr Department of Electrical and Computer Engineering, George Mason University\\
       Fairfax, VA 22030, USA
}

\editor{Corinna Cortes}
\maketitle

\begin{abstract} 
This paper studies the decentralized optimization and learning problem where multiple interconnected agents aim to learn an optimal decision function defined over a reproducing kernel Hilbert space by jointly minimizing a global objective function, with access to their own locally observed dataset. As a non-parametric approach, kernel learning faces a major challenge in distributed implementation: the decision variables of local objective functions are data-dependent and thus cannot be optimized under the decentralized consensus framework without any raw data exchange among agents. To circumvent this major challenge, we leverage the random feature (RF) approximation approach to enable consensus on the function modeled in the RF space by data-independent parameters across different agents. We then design an iterative algorithm, termed DKLA, for fast-convergent implementation via ADMM. Based on DKLA, we further develop a communication-censored kernel learning (COKE) algorithm that reduces the communication load of DKLA by preventing an agent from transmitting at every iteration unless its local updates are deemed informative. Theoretical results in terms of linear convergence guarantee and generalization performance analysis of DKLA and COKE are provided. Comprehensive tests on both synthetic and real datasets are conducted to verify the communication efficiency and learning effectiveness of COKE.\footnote{Preliminary results in this paper were presented in part at the 2019 IEEE Data Science Workshop~\citep{xu2019coke}.}
\end{abstract}
\begin{keywords}
Decentralized nonparametric learning, reproducing kernel Hilbert space, random features, ADMM, communication censoring.
\end{keywords}

\section{Introduction}
Decentralized learning has attracted extensive interest in recent years, largely due to the explosion of data generated everyday from mobile sensors, social media services, and other networked multi-agent applications~\citep{worden2006application, ilyas2013distributed, facchinei2015parallel, demarie2019machine}. In many of these applications, the observed data are usually kept private at local sites without being aggregated to a fusion center, either due to the prohibitively high cost of raw data transmission or privacy concerns. Meanwhile, each agent in the network only communicates with its one-hop neighbors within its local area to save transmission power. Such localized data processing and transmission obviate the implementation of any centralized learning techniques. Under this circumstance, this article focuses on the decentralized learning problem where a network of distributed agents aim to collaboratively learn a functional model describing the global data with only access to their own locally observed datasets.

To learn the functional model that is often nonlinear and complex, nonparametric kernel methods are widely appreciated thanks to the ``kernel trick" that makes some well-behaved linear learning algorithms applicable in a high-dimensional implicit feature space, without explicit mapping from data to that feature space~\citep{shawe2004kernel, hofmann2008kernel, perez2004kernel}. However, in the absence of any raw data sharing or aggregation, it is challenging to directly apply them to a decentralized multi-agent setting and solve them under the consensus optimization framework using algorithms such as decentralized alternating direction method of multipliers (ADMM)~\citep{shi2014linear}. This is because decentralized learning relies on solving local optimization problems and then aggregating the updates on the local decision variables over the network through one-hop communications in an iterative manner~\citep{nedic2016tutorial}. Unfortunately, these decision variables of local objective functions resulted from the kernel trick are data-dependent and thus cannot be optimized in the absence of raw data exchange under the decentralized consensus framework.  

There are several works applying kernel methods in decentralized learning for various applications under different settings~\citep{predd2006distributed, mitra2014diffusion, gao2015diffusion, chouvardas2016diffusion, shin2016distributed, shin2018distributed, koppel2018decentralized}. These works, however, either assume that agents have access to their neighbors' observed raw data~\citep{predd2006distributed} or require agents to transmit their raw data to their neighbors~\citep{koppel2018decentralized} to ensure consensus through collaborative learning. These assumptions may not be valid in many practical applications that involve users' private data. Moreover, standard kernel learning for big data faces the curse of dimensionality when the number of training examples increases~\citep{shawe2004kernel}. For example, in~\citep{mitra2014diffusion, chouvardas2016diffusion}, the nonlinear function learned at each node is represented as a weighted combination of kernel functions centered on its local observed data. As a result, each agent needs to transmit both the weights of kernel functions and its local data to its neighbors at every iterative step to guarantee consensus of the common prediction function. Thus, both the computation and communication resources are demanding in the distributed implementation. To alleviate the curse of dimensionality problem, \citet{gao2015diffusion} and \citet{koppel2018decentralized} have developed compression techniques such as data selection and sparse subspace projection, respectively, but these techniques typically incur considerable extra computation, and still involve raw data exchange with no alleviation to the data privacy concern. Furthermore, when computation cost is more affordable than the communication in the big data scenario, communication cost of the iterative learning algorithms becomes the bottleneck for efficient distributed learning~\citep{mcmahan2016communication}. Therefore, it is crucial to design communication-efficient distributed kernel learning algorithms with data privacy protection.

\subsection{Related work}
This work lies at the intersection of non-parametric kernel methods, decentralized learning with batch-form data, and communication-efficient iterative implementation. Related work to these three subjects is reviewed below.  

\noindent
\textbf{Centralized kernel learning.} Centralized kernel methods assume data are collected and processed by a single server and are known to suffer from the curse of dimensionality for large-scale learning tasks. To mitigate their computational complexity, various dimensionality reduction techniques are developed for both batch-form or online streaming learning, including stochastic approximation~\citep{bucak2010multi, gu2018asynchronous}, restricting the number of function parameters~\citep{gomes2010budgeted, wang2012breaking, zhang2013online, le2016nonparametric, koppel2017parsimonious}, and approximating the kernel during training~\citep{honeine2015analyzing, engel2004kernel,richard2008online, drineas2005nystrom, dai2014scalable, lu2016large, sheikholeslami2018large, rahimi2008random, buazuavan2012fourier, nguyen2017large}. Among them, random feature (RF) mapping methods have gained popularity thanks to their ability to map the large-scale data into a RF space of much reduced dimension by approximating the kernel with a fixed (small) number of random features, which thus circumvents the curse of dimensionality problem~\citep{rahimi2008random, dai2014scalable, buazuavan2012fourier, nguyen2017large}. Enforcing orthogonality on random features can greatly reduce the error in kernel approximation~\citep{yu2016orthogonal, shen2018online}, and the learning performance of RF-based methods is evaluated in~\citep{bach2017equivalence, rudi2017generalization, li2018towards}.

\noindent
\textbf{Decentralized kernel learning.} For the decentralized kernel learning problem relevant to our work~\citep{mitra2014diffusion, gao2015diffusion, chouvardas2016diffusion, koppel2018decentralized}, gradient descent is conducted locally at each agent to update its learning model, followed by diffusion-based information exchange among agents. However, these methods either assume that agents have access to their neighbors' observed raw data or require agents to transmit their raw data to their neighbors to ensure convergence on the prediction function. For the problem studied in this article where the observed data are only locally available, these methods are not applicable since there are no common decision parameters for consensus without any raw data exchange. Moreover, these methods operate in the kernel space parameterized by training data, and still encounter the curse of dimensionality when the local dataset goes large. Though data selection~\citep{gao2015diffusion} and subspace projection~\citep{koppel2018decentralized} are adopted to alleviate the curse of dimensionality problem, they typically require significant extra computational resources. RF mapping~\citep{rahimi2008random} offers a viable approach to overcome these issues, by having all agents map their datasets of various sizes onto the same RF space. For instance,~\citet{bouboulis2018online} proposes a diffusion-based combine-then-adapt (CTA) method that achieves consensus on the model parameters in the RF space for the online learning problem, without the exchange of raw data. Though the batch-form counterpart of online CTA can be developed for off-line learning, the convergence speed of the diffusion-based method is relatively slow compared with higher-order methods such as ADMM~\citep{liu2019communication}.  	
	
\noindent
\textbf{Communication-efficient optimization.} Communication-efficient algorithms for decentralized optimization and learning problems have attracted attention when data movement among computing nodes becomes a bottleneck due to the high latency and limited bandwidth of decentralized networks. To reduce the communication cost, one way is to transmit the compressed information by quantization~\citep{zhu2016quantized, NIPS2017_6768, zhang2019quantized} or sparsification~\citep{NIPS2018_7697, NIPS2018_7837, NIPS2018_7405, harrane2018reducing}. However, these methods only reduce the required bandwidth at each communication round, not the number of rounds or the number of transmissions. Alternatively, some works randomly select a number of nodes for broadcasting/communication and operate asynchronous updating to reduce the number of transmissions per iteration~\citep{mota2013d, li2014communication, jaggi2014communication, arablouei2015analysis, mcmahan2016communication, yin2018communication, NIPS2019_8694}. In contrast to random node selection, a more intuitive way is to evaluate the importance of a message in order to avoid unnecessary transmissions~\citep{chen2018lag, liu2019communication, li2019cola}. This is usually implemented by adopting a censoring scheme to adaptively decide if a message is informative enough to be transmitted during the iterative optimization process. Other efforts to improve the communication efficiency are made by accelerating the convergence speed of the iterative algorithm implementation~\citep{shamir2014Comm, reddi2016aide, li2019communication}.

\subsection{Contributions}
This paper develops communication-efficient decentralized kernel learning algorithms under the consensus optimization framework without any central coordination or raw data exchange among agents for built-in privacy protection. Relative to prior art, our contributions are summarized as follows.
\begin{itemize}
\item We first formulate the decentralized multi-agent kernel learning problem as a decentralized consensus optimization problem in the RF space. Since most machine learning scenarios can afford plenty computational capability but limited communication resources, we solve this problem with ADMM, which has shown fast convergence at the expense of relatively high computation cost per iteration~\citep{shi2014linear}. To the best of our knowledge, this is the first work to solve decentralized kernel learning in the RF space by ADMM without any raw data exchange. The key of our proposed Decentralized Kernel Learning via ADMM (DKLA) algorithm is to apply RF mapping, which not only reduces the computational complexity but also enables consensus on a set of model parameters of fixed size in the RF space. In addition, since no raw data is exchanged among agents and the mapping from the original data space to the RF space is not one-to-one mapping, data privacy is protected to a certain level.

\item To increase the communication efficiency, we further develop a COmmunication-censored KErnel learning (COKE) algorithm, which achieves desired learning performance given limited communication resources and energy supply. Specifically, we devise a simple yet powerful censoring strategy to allow each user to autonomously skip unnecessary communications when its local update is not informative enough for transmission, without aid of a central coordinator. In this way, the communication efficiency can be boosted at almost no sacrifice of the learning performance. When the censoring strategy is absent, COKE degenerates to DKLA. 

\item In addition, we conduct theoretical analysis in terms of both functional convergence and generalization performance to provide guidelines for practical implementations of our proposed algorithms. We show that the individually learned functional at each agent through DKLA and COKE both converges to the optimal one at a linear rate under mild conditions. For the generalization performance, we show that $O(\sqrt{T}\log d_{\mathbf{K}}^\lambda)$ features are sufficient to ensure $O(1/\sqrt{T})$ learning risk for the decentralized kernel ridge regression problem, where $d_{\mathbf{K}}^\lambda$ is the number of effective degrees of freedom that will be defined in Section 4.2 and $T$ is the total number of samples.

\item Finally, we test the performance of our proposed DKLA and COKE algorithms on both synthetic and real datasets. The results corroborate that both DKLA and COKE exhibit attractive learning performance and COKE is highly communication-efficient. 
\end{itemize}

\subsection{Organization and notation of the paper}
\textbf{Organization.} Section \ref{sec:prosta} formulates the problem of non-parametric learning and highlights the challenges in applying traditional kernel methods in the decentralized setting. Section \ref{sec:alog_develop} develops the decentralized kernel learning algorithms, including both DKLA and COKE. Section \ref{sec:analytic} presents the theoretical results and Section \ref{sec:exper} reports the numerical tests using both synthetic data and real datasets. Concluding remarks are provided in Section \ref{sec:con}.

\noindent
\textbf{Notation.} $\mathbb{R}$ denotes the set of real numbers. $\|\cdot\|_2$ denotes the Euclidean norm of vectors and $\|\cdot\|_F$ denotes the Frobenius norm of matrices. $|\cdot|$ denotes the cardinality of a set. $\mathbf{A}$, $\mathbf{a}$, and $a$ denotes a matrix, a vector, and a scalar, respectively.

\section{Problem Statement}
\label{sec:prosta}
This section reviews basics of kernel-based learning and decentralized optimization, introduces notation, and provides background needed for our novel DKLA and COKE schemes.

Consider a network of $N$ agents interconnected over a fixed topology $\mathcal{G}=(\mathcal{N}, \mathcal{C}, \mathbf{A})$, where $\mathcal{N}=\{1,2 \dots, N\}$, $ \mathcal{C} \subseteq\mathcal{N}\times\mathcal{N}$, and $\mathbf{A}\in \mathbb{R}^{N\times N}$ denote the agent set, the edge set and the adjacency matrix, respectively. The elements of $\mathbf{A}$ are $a_{in}=a_{ni}=1$ when the unordered pair of distinct agents $(i,n)\in \mathcal{C}$, and $a_{in}=a_{ni}=0$ otherwise. For agent $i$, its one-hop neighbors are in the set $\mathcal{N}_i=\{n|(n,i)\in \mathcal{C}\}$. The term \textit{agent} used here can be a single computational system (e.g. a smart phone, a database, etc.) or a collection of co-located computational systems (e.g. data centers, computer clusters, etc.). Each agent only has access to its locally observed data composed of independently and identically distributed (i.i.d) input-label pairs $\{\mathbf{x}_{i,t},y_{i,t}\}_{t=1}^{T_i}$ obeying an unknown probability distribution $p$ on $\mathcal{X}\times \mathcal{Y}$, with $\mathbf{x}_{i,t} \in \mathbb{R}^d$ and $y_{i,t} \in \mathbb{R}$. The kernel learning task is to find a prediction function $f$ that best describes the ensemble of all data from all agents. Suppose that $f$ belongs to the reproducing kernel Hilbert space (RKHS) $\mathcal{H}:=\{f|f(\mathbf{x})=\sum_{t=1}^\infty \alpha_t \kappa(\mathbf{x},\mathbf{x}_t)\}$ induced by a positive semidefinite kernel $\kappa(\mathbf{x},\mathbf{x}_t):\mathbb{R}^d\times\mathbb{R}^d\rightarrow \mathbb{R}$ that measures the similarity between $\mathbf{x}$ and $\mathbf{x}_t$, for all $\mathbf{x}, \mathbf{x}_t\in\mathcal{X} $. In a decentralized setting with privacy concern, this means that each agent has to be able to learn the global function $f\in\mathcal{H}$ such that $y_{i,t}=f(\mathbf{x}_{i,t})+e_{i,t}$ for $\{\{\mathbf{x}_{i,t},y_{i,t}\}_{t=1}^{T_i}\}_{i=1}^N$, without exchange of any raw data and in the absence of a fusion center, where the error terms $e_{i,t}$ are minimized according to certain optimality metric.

To evaluate the learning performance, a nonnegative loss function $\ell(y,\hat{y})$ is utilized to measure the difference between the true label value $y$ and the predicted value $\hat{y}=f(\mathbf{x})$. Some common loss functions include the quadratic loss $\ell(y,\hat{y})=(y-\hat{y})^2$ for regression tasks, the hinge loss $\ell(y,\hat{y})=\max(0,1-y\hat{y})$ and the logistic loss $\ell(y,\hat{y})=\log(1+e^{-y\hat{y}})$ for binary classification tasks. The above mentioned loss functions are all convex with respect to $\hat{y}$. The learning problem is then to minimize the expected risk of the prediction function:
\begin{equation}
\label{eq:expected_risk}
R(f)=\int_{\mathcal{X}\times \mathcal{Y} }  \ell(f(\mathbf{x}),y) dp(\mathbf{x},y),  
\end{equation}  
which indicates the generalization ability of $f$ to new data. 

However, the distribution $p$ is unknown in most learning tasks. Therefore, minimizing $R(f)$ is not applicable. Instead, given the finite number of training examples, the problem turns to minimizing the empirical risk:
\begin{equation}
\label{eq:empirical_risk}
\underset{f\in \mathcal{H}}{\text{min}} \quad  \hat{R}(f):=\sum_{i=1}^N \hat{R}_i(f),
\end{equation}
where $\hat{R}_i(f)$ is the local empirical risk for agent $i$ given by 
\begin{equation}
\label{eq:empirical_risk_f_i} 
\hat{R}_i(f) =\frac{1}{T_i}\sum_{t=1}^{T_i}\ell(f(\mathbf{x}_{i,t}),y_{i,t})+\lambda_i \|f\|_{\mathcal{H}}^2, 
\end{equation}
with $\|\cdot\|_{\mathcal{H}}$ being the norm associated with $\mathcal{H}$, and $\lambda_i>0$ being a regularization parameter that controls over-fitting.

The representer theorem states that the minimizer of a regularized empirical risk functional defined over a RKHS can be represented as a finite linear combination of kernel functions evaluated on the data pairs from the training dataset~\citep{scholkopf2001generalized}. If $\{\{\mathbf{x}_{i,t},y_{i,t}\}_{t=1}^{T_i}\}_{i=1}^N$ are centrally available at a fusion center, the minimizer of~\eqref{eq:empirical_risk} admits
\begin{equation}
\label{eq:minimizer_alpha}
f^\star(\mathbf{x})=\sum_{i=1}^N\sum_{t=1}^{T_i} \alpha_{i,t} \kappa(\mathbf{x},\mathbf{x}_{i,t}):=\bm{\alpha}^\top\bm{\kappa}(\mathbf{x}),
\end{equation}
where $\bm{\alpha}=[\alpha_{1,1},\dots,\alpha_{N,T_N}]^\top\in \mathbb{R}^T$ is the coefficient vector to be learned, $T=\sum_{i=1}^N T_i$ is the total number of samples, and $\bm{\kappa}(\mathbf{x}) = [\kappa(\mathbf{x},\mathbf{x}_{1,1}),\dots,\kappa(\mathbf{x},\mathbf{x}_{N,T_N})]^\top\in\mathbb{R}^T$ is the kernel function parameterized by the global data $\mathbf{X}_T: = \{\{\mathbf{x}_{i,t}\}_{t=1}^{T_i}\}_{i=1}^N$ from all agents, for any $\mathbf{x}$. In RKHS, since $\langle \kappa(\mathbf{x}_t,\mathbf{x}), \kappa(\mathbf{x}_\tau,\mathbf{x}) \rangle_\mathcal{H} = \kappa(\mathbf{x}_t,\mathbf{x}_\tau)$, it yields $\|f\|_{\mathcal{H}}^2=\bm{\alpha}^\top\mathbf{K}\bm{\alpha}$, where $\mathbf{K}$ is the $T\times T$ kernel matrix that measures the similarity between any two data points in $\mathbf{X}_T$. 
In this way, the local empirical risk~\eqref{eq:empirical_risk_f_i} can be reformulated as a function of $\bm{\alpha}$:
\begin{equation}
\label{eq:empirical_risk_alpha_i}
\begin{split}
\hat{R}_i(\bm{\alpha}):&=\frac{1}{T_i}\sum_{t=1}^{T_i}\ell(f^\star(\mathbf{x}_{i,t}),y_{i,t})+ \lambda_i \|f^\star\|_{\mathcal{H}}^2= \frac{1}{T_i} \sum_{t=1}^{T_i} \ell(\bm{\alpha}^\top\bm{\kappa}(\mathbf{x}_{i,t}),y_{i,t})+\lambda_i\bm{\alpha}^\top \mathbf{K}\bm{\alpha}.
\end{split}
\end{equation}
Accordingly, \eqref{eq:empirical_risk} becomes
\begin{equation}
\label{eq:empirical_alpha_N}
\underset{\bm{\alpha}\in \mathbb{R}^{T}}{\text{min}} \quad \sum_{i=1}^N \hat{R}_i(\bm{\alpha}). 
\end{equation}

Relating the decentralized kernel learning problem with the decentralized consensus optimization problem, solving~\eqref{eq:empirical_alpha_N} is equivalent to solving 
\begin{equation}
\label{eq:empirical_risk_dis_f}
\begin{split}
\underset{\{\bm{\alpha}_i\in \mathbb{R}^{T}\}_{i=1}^N}{\text{min}} \quad &\sum_{i=1}^N \hat{R}_i(\bm{\alpha}_i)\\
\text{s.t.} \qquad  &\bm{\alpha}_i = \bm{\alpha}_n, \qquad \forall i,\quad \forall n\in \mathcal{N}_i, 
\end{split}
\end{equation} 
where $\bm{\alpha}_i$ and $\bm{\alpha}_n$ are the local copies of the global decision variable $\bm{\alpha}$ at agent $i$ and agent $n$, respectively. The problem can then be solved by ADMM~\citep{shi2014linear} or other primal dual methods~\citep{terelius2011decentralized}. However, it is worth noting that~\eqref{eq:empirical_risk_dis_f} reveals a subtle yet profound difference from a general optimization problem for parametric learning. That is, each local function $\hat{R}_i$ depends on not only the global decision variable $\bm{\alpha}$, but also the global data $\mathbf{X}_T$ because of the kernel terms $\bm{\kappa}(\mathbf{x}_{i,t})$ and $\mathbf{K}$. As a result, solving the local objective for agent $i$ requires raw data from all other agents to obtain $\bm{\kappa}(\mathbf{x}_{i,t})$ and $\mathbf{K}$, which contradicts the situation that private raw data are only locally available. Moreover, notice that $\bm{\alpha}_i$ is of the same size $T$ as that of the ensemble dataset, which incurs the curse of dimensionality and insurmountable computational cost when $T$ becomes large, even when the obstacle of making all the data available to all agents is not of concern.

To resolve this issue, an alternative formulation is to associate a local prediction model $\bar{f}_i\in \mathcal{H}$ with each agent $i$, with $\bar{f}_i^\star=\sum_{t=1}^{T_i} \bar{\alpha}_{i,t} \kappa(\mathbf{x},\mathbf{x}_{i,t})=\bar{\bm{\alpha}}_i^\top\bm{\kappa}_i(\mathbf{x})$ being the local optimal solution that only involves local data $ \{\mathbf{x}_{i,t}\}_{t=1}^{T_i}$~\citep{ji2016distributed}. Specifically, $\bar{\bm{\alpha}}_i=[\bar{\alpha}_{i,1},\dots,\bar{\alpha}_{i,T_i}]\in \mathbb{R}^{T_i}$, and $\bm{\kappa}_i(\mathbf{x}) = [\kappa(\mathbf{x},\mathbf{x}_{i,1}),\dots,\kappa(\mathbf{x},\mathbf{x}_{i,T_i})]^\top\in\mathbb{R}^{T_i}$ is parameterized by the local data $\{\mathbf{x}_{i,t}\}_{t=1}^{T_i}$ only. In this way, the local cost function becomes
\begin{equation}
\label{eq:empirical_risk_alpha_i_alter}
\begin{split}
\hat{R}_i(\bar{\bm{\alpha}}_i):&=\frac{1}{T_i}\sum_{t=1}^{T_i}\ell(\bar{f}^\star_i(\mathbf{x}_{i,t}),y_{i,t})+\lambda_i \|\bar{f}_i^\star\|_{\mathcal{H}}^2 = \frac{1}{T_i} \sum_{t=1}^{T_i} \ell(\bar{\bm{\alpha}}_i^\top\bm{\kappa}_i(\mathbf{x}_{i,t}),y_{i,t})+\lambda_i \bar{\bm{\alpha}}_i^\top \mathbf{K}_i\bar{{\bm{\alpha}}}_i,
\end{split}
\end{equation}
where $\mathbf{K}_i$ is of size $T_i\times T_i$ and depends on local data only. With \eqref{eq:empirical_risk_alpha_i_alter}, the optimization problem \eqref{eq:empirical_risk_dis_f} is then modified to 
\begin{equation}
\label{eq:empirical_risk_dis_alpha}
\begin{split}
\underset{\{\bar{\bm{\alpha}}_i\in \mathbb{R}^{T_i}\}_{i=1}^N}{\text{min}} \quad &\sum_{i=1}^N \hat{R}_i(\bar{\bm{\alpha}}_i)\\
\text{s.t.} \qquad  &\bar{f}_n(\mathbf{x}_{i,t}) = \bar{f}_i(\mathbf{x}_{i,t}), \qquad \forall i,\quad \forall n\in \mathcal{N}_i, \quad t=1,\dots, T_i, 
\end{split}
\end{equation} 
and can be solved distributedly by ADMM. Note that the consensus constraint is the learned prediction values $\bar{f}_i(\mathbf{x})$, not the parameters $\bar{\bm{\alpha}}_i$. This is because $\bar{\bm{\alpha}}_i$ are data-dependent and may have different sizes at different agents (the dimension of $\bar{\bm{\alpha}}_i$ equals to the number of training samples at agent $i$), and cannot be directly optimized through consensus.

Still, this method has four drawbacks. Firstly, it is necessary to associate a local learning model $\bar{f}_i$ to each agent $i$ for the decentralized implementation. However, the local learning model $\bar{f}_i$ and the global optimal model $f$ in \eqref{eq:empirical_risk} may not be the same because different local training data are used. Therefore, the optimization problem \eqref{eq:empirical_risk_dis_alpha} is only an approximation of \eqref{eq:empirical_risk}. Even with the equality constraint to minimize the gap between the decentralized learning output and the optimal centralized one, the approximation performance is not guaranteed. Besides, the functional consensus constraint still requires raw data exchange among agents in order for agent $n\in \mathcal{N}_i$ to be able to compute the values $\bar{f}_n(\mathbf{x}_{i,t})$ from agent $i$'s data $\mathbf{x}_{i,t}$, for $i\neq n$. Apparently, this violates the privacy-protection requirement for practical applications. In addition, when $T_i$ is large, both the storage and computational costs are high for each agent due to the curse of dimensionality problem at the local sites. Lastly, the frequent local communication is resource-consuming under communication constraints. To circumvent all these obstacles, the goal of this paper is to develop efficient decentralized algorithms that protect privacy and conserve communication resources.

\section{Algorithm Development}
\label{sec:alog_develop}
In this section, we leverage the RF approximation and ADMM to develop our algorithms. We first introduce the RF mapping method. Then, we devise the DKLA algorithm that globally optimizes a shared learning model for the multi-agent system. Finally, we take into consideration of the limited communication resources in large-scale decentralized networks and develop the COKE algorithm. Both DKLA and COKE are computationally efficient and protect data privacy at the same time. Further, COKE is communication efficient.

\subsection{RF-based kernel learning}
\label{subsec:RF_kernel}
As stated in previous sections, standard kernel methods incur the curse of dimensionality issue when the data size grows large. To make kernel methods scalable for a large dataset, RF mapping is adopted for approximation by using the shift-invariance property of kernel functions~\citep{rahimi2008random}. 

For a shift-invariant kernel that satisfies $\kappa(\mathbf{x}_{t},\mathbf{x}_\tau) = \kappa(\mathbf{x}_{t}-\mathbf{x}_\tau),\;\forall t,\;\forall\tau$, if $\kappa(\mathbf{x}_{t}-\mathbf{x}_\tau)$ is absolutely integrable, then its Fourier transform $p_\kappa(\bm{\omega})$ is guaranteed to be nonnegative ($p_\kappa(\bm{\omega})\geq 0$), and hence can be viewed as its probability density function (pdf) when $\kappa$ is scaled to satisfy $\kappa(0)=1$~\citep{bochner2005harmonic}. Therefore, we have  
\begin{equation}
\label{eq:kern_fouri}
\kappa(\mathbf{x}_{t},\mathbf{x}_\tau) = \int  p_\kappa(\bm{\omega} )e^{j\bm{\omega}^\top(\mathbf{x}_{t}-\mathbf{x}_\tau)}d\bm{\omega} :=\mathbb{E}_{\bm{\omega}}[e^{j\bm{\omega}^\top(\mathbf{x}_{t}-\mathbf{x}_\tau)}]= \mathbb{E}_{\bm{\omega}}[\phi(\mathbf{x}_{t},\bm{\omega})\phi^\ast(\mathbf{x}_{\tau},\bm{\omega})],
\end{equation} 
where $\mathbb{E}$ denotes the expectation operator, $\phi(\mathbf{x},\bm{\omega}):= e^{j\bm{\omega}^\top \mathbf{x}}$ with $\bm{\omega}\in\mathbb{R}^d$, and $\ast$ is the complex conjugate operator. In \eqref{eq:kern_fouri}, the first equality is the result of the Fourier inversion theorem, and the second equality arises by viewing $p_{\kappa}(\bm{\omega})$ as the pdf of $\bm{\omega}$. In this paper, we adopt a Gaussian kernel $\kappa(\mathbf{x}_{t},\mathbf{x}_\tau) =\rm{exp}(-\|\mathbf{x}_{t}-\mathbf{x}_{\tau}\|_2^2/(2\sigma^2))$, whose pdf is a normal distribution with $p_\kappa(\bm{\omega})\sim\mathbf{N}(\mathbf{0}, \sigma^{-2} \mathbf{I})$.  

The main idea of RF mapping is to randomly generate $\{\bm{\omega}_l\}_{l=1}^L$ from the distribution $p_\kappa(\bm{\omega})$ and approximate the kernel function $\kappa(\mathbf{x}_{t},\mathbf{x}_\tau)$ by the sample average
\begin{equation}
\label{eq:kernel_map_L}
\hat{\kappa}_L(\mathbf{x}_{t},\mathbf{x}_\tau):=\frac{1}{L}\sum_{l=1}^L
\phi(\mathbf{x}_t,\bm{\omega}_l)\phi^\ast(\mathbf{x}_\tau,\bm{\omega}_l):=\bm{\phi}_L^\dagger (\mathbf{\mathbf{x}}_{\tau})\bm{\phi}_L(\mathbf{\mathbf{x}}_{t}),
\end{equation}
where $\bm{\phi}_L(\mathbf{x}):=\sqrt{\frac{1}{L}}[\phi(\mathbf{x},\bm{\omega}_1),\dots, \phi(\mathbf{x},\bm{\omega}_L)]^\top$ and $\dagger$ is the conjugate transpose operator.

The following real-valued mappings can be adopted to approximate $\kappa(\mathbf{x}_{t},\mathbf{x}_\tau)$, both satisfying the condition $\mathbb{E}_{\bm{\omega}}[\phi_r(\mathbf{x}_{t},\bm{\omega})^\top\phi_r(\mathbf{x}_{\tau},\bm{\omega})]=\kappa(\mathbf{x}_{t},\mathbf{x}_{\tau})$~\citep{rahimi2008random}: 
\begin{align}
\phi_r(\mathbf{x},\bm{\omega})&=[\cos(\bm{\omega} ^\top\mathbf{x}),\sin(\bm{\omega}^\top\mathbf{x})]^\top,\label{eq:realRF1}\\
\phi_r(\mathbf{x},\bm{\omega})&=\sqrt{2}\cos(\bm{\omega}^\top \mathbf{x}+b)\label{eq:realRF2},
\end{align}
where $b$ is drawn uniformly from $[0,2\pi]$.

With the real-valued RF mapping, the minimizer of \eqref{eq:empirical_risk} then admits the following form:
\begin{equation}
\label{eq:minimizer_RF}
\hat{f}^\star(\mathbf{x})=\sum_{i=1}^N\sum_{t=1}^{T_i}\alpha_{i,t}\bm{\phi}_L^\top(\mathbf{x}_{i,t})\bm{\phi}_L(\mathbf{x})=\bm{\theta}^\top\bm{\phi}_L(\mathbf{x}),
\end{equation}
where $\bm{\theta}^\top:=\sum_{i=1}^N\sum_{t=1}^{T_i}\alpha_{i,t}\bm{\phi}_L^\top(\mathbf{x}_{i,t})$ denotes the new decision vector to be learned in the RF space and $\bm{\phi}_L(\mathbf{x}) = \sqrt{\frac{1}{L}} [\phi_r(\mathbf{x},\bm{\omega}_1),\dots,\phi_r(\mathbf{x},\bm{\omega}_L)]^\top$. If \eqref{eq:realRF1} is adopted, then $\bm{\phi}_L(\mathbf{x})$ and $\bm{\theta}$ are of size $2L$. Otherwise, if \eqref{eq:realRF2} is adopted, then $\bm{\phi}_L(\mathbf{x})$ and $\bm{\theta}$ are of size $L$. In either case, the size of $\bm{\theta}$ is fixed and does not increase with the number of data samples.

\subsection{DKLA: Decentralized kernel learning via ADMM }
\label{subsec:RF_Dectr}
Consider the decentralized kernel learning problem described in Section \ref{sec:prosta} and adopt the RF mapping described in Section \ref{subsec:RF_kernel}. Let all agents in the network have the same set of random features, i.e., $\{\bm{\omega}_l\}_{l=1}^L$. 
Plugging~\eqref{eq:minimizer_RF} into the local cost function $\hat{R}_i(f)$ in \eqref{eq:empirical_risk_f_i} gives
\begin{equation}
\label{eq:empirical_risk_theta_i}
\begin{split}
\hat{R}_i(\bm{\theta}):&=\frac{1}{T_i}\sum_{t=1}^{T_i}\ell(\hat{f}^\star(\mathbf{x}_{i,t}),y_{i,t})+\lambda_i\|\hat{f}^\star\|_{\mathcal{H}}^2= \frac{1}{T_i}\sum_{t=1}^{T_i}\ell(\bm{\theta}^\top\bm{\phi}_L(\mathbf{x}_{i,t}), y_{i,t})+\lambda_i\|\bm{\theta}\|_2^2.
\end{split}
\end{equation}
In \eqref{eq:empirical_risk_theta_i}, we have
\begin{equation}
\begin{split}
\|\bm{\theta}\|_2^2:&=(\sum_{i=1}^N\sum_{t=1}^{T_i} \alpha_{i,t} \bm{\phi}_L^\top(\mathbf{x}_{i,t}))( \sum_{n=1}^N\sum_{\tau=1}^{T_i}\alpha_{n,\tau}\bm{\phi}_L (\mathbf{x}_{n,\tau})) \\
& =\sum_{i=1}^N\sum_{t=1}^{T_i} \sum_{n=1}^N\sum_{\tau=1}^{T_i}  \alpha_{i,t} \alpha_{n,\tau} \kappa(\mathbf{x}_{i,t},\mathbf{x}_{n,\tau}):=\|\hat{f}^\star\|_{\mathcal{H}}^2\nonumber.
\end{split}
\end{equation}
 
Therefore, with RF mapping, the centralized benchmark \eqref{eq:empirical_risk} becomes
\begin{equation}
\label{eq:empirical_risk_theta_N}
\underset{\bm{\theta}\in \mathbb{R}^{L}}{\text{min}} \quad \sum_{i=1}^N \hat{R}_i(\bm{\theta}).
\end{equation} 
Here for notation simplicity, we denote the size of $\bm{\theta}$ by $L\times 1$, which can be achieved by adopting the real-valued mapping in~\eqref{eq:realRF2}. Adopting an alternative mapping such as~\eqref{eq:realRF1} only changes the size of $\bm{\theta}$ while the algorithm development is the same. RF mapping is essential because it results in a common optimization parameter $\bm{\theta}$ of fixed size for all agents.

To solve \eqref{eq:empirical_risk_theta_N} in a decentralized manner, we associate a model parameter $\bm{\theta}_i$ with each agent $i$ and enforce the consensus constraint on neighboring agents $i$ and $n$ using an auxiliary variable $\bm{\vartheta}_{in}$. Specifically, the RF-based decentralized kernel learning problem is formulated to jointly minimize the following objective function:
\begin{equation}
\label{eq:empirical_risk_dis_theta}
\begin{split}
\underset{\{\bm{\theta}_i\in \mathbb{R}^{L}\}, \{\bm{\vartheta}_{in}\in \mathbb{R}^{L}\} }{\text{min}} \quad &\sum_{i=1}^N \hat{R}_i(\bm{\theta}_i)\\
\text{s.t.} \qquad \quad &\bm{\theta}_i=\bm{\vartheta}_{in},\; \bm{\theta}_n=\bm{\vartheta}_{in}, \qquad \forall (i,n)\in \mathcal{C}.
\end{split}
\end{equation} 

Note that the new decision variables $\bm{\theta}_i$ to be optimized are local copies of the global optimization parameter $\bm{\theta}$ and are of the same size for all agents. On the contrary, the decision variables $\bar{\bm{\alpha}}_i$ in~\eqref{eq:empirical_risk_dis_alpha} are data-dependent and may have different sizes. In addition, the size of $\bm{\theta}$ is $L$, which can be much smaller than that of $\bm{\alpha}$ (whose size equals to $T$) in~\eqref{eq:empirical_alpha_N}. For big data scenarios where $L\ll T$, RF mapping greatly reduces the computational complexity. Moreover, as shown in the following, the updating of $\bm{\theta}$ does not involve any raw data exchange and the RF mapping from $\mathbf{x}$ to $\bm{\phi}_L (\mathbf{x})$ is not one-to-one mapping, therefore provides raw data privacy protection. Further, it is easy to set the regularization parameters $\lambda_i$ to control over-fitting. Specifically, since the parameters $\bm{\theta}_i$ are of the same length among agents, we can set them to be $\lambda_i=\frac{1}{N}\lambda,\forall i$, where $\lambda$ is the corresponding over-fitting control parameter assuming all data are collected at a center. In contrast, the regularization parameters $\lambda_i$ in~\eqref{eq:empirical_risk_alpha_i} depend on local data and need to satisfy $\lambda = \sum_{i=1}^N \lambda_i$, which is relatively difficult to tune in a large-scale network.

In the constraint, $\bm{\theta}_i$ are separable when $\bm{\vartheta}_{in}$ are fixed, and vice versa. Therefore, \eqref{eq:empirical_risk_dis_theta} can be solved by ADMM. Following~\citep{shi2014linear}, we develop the DKLA algorithm where each agent updates its local primal variable $\bm{\theta}_i$ and local dual variable $\bm{\gamma}_i$ by
 
\begin{subequations} 
	\label{eq:ADMM_update}
	\begin{align} 
	\textstyle
	&\bm{\theta}_i^k := \arg \min_{\bm{\theta}_i} \;\left \{\hat{R}_i(\bm{\theta}_i)+\rho|\mathcal{N}_i|\|\bm{\theta}_i\|_2^2 + \bm{\theta}_i^\top\left[\bm{\gamma}_i^{k-1}-\rho\sum_{n\in\mathcal{N}_i}\left(\bm{\theta}_i^{k-1}+ \bm{\theta}_n^{k-1}\right)\right]\right\},\label{eq:ADMM_update_a}  \\
	&\bm{\gamma}_i^k =\bm{\gamma}_i^{k-1}+\rho\sum_{n\in\mathcal{N}_i}\Big(\bm{\theta}_i^k-\bm{\theta}_n^k\Big)\label{eq:ADMM_update_b},
	\end{align} 
\end{subequations}
where $|\mathcal{N}_i|$ is the cardinality of $\mathcal{N}_i$. The auxiliary variable $\bm{\vartheta}_{in}$ can be written as a function of $\bm{\theta}_{i}$ and then canceled out. Interested readers are referred to~\citep{shi2014linear} for detailed derivation. The learning algorithm DKLA is outlined in Algorithm \ref{tab:dis_RF_ADMM}. Note that the random features need to be common to all agents, hence, in step 1, we restrict them to be drawn according to a common random seed. Algorithm \ref{tab:dis_RF_ADMM} is fully decentralized since the updates of $\bm{\theta}_i$ and $\bm{\gamma}_i$ depend only on local and neighboring information.
 
\begin{algorithm}[t]
	\caption{DKLA Run at Agent $i$ }
	\label{tab:dis_RF_ADMM}
	\begin{algorithmic}[1]
		\REQUIRE Kernel $\kappa$, the number of random features $L$, and $\lambda$ to control over-fitting; initialize local variables to $\bm{\theta}_i^0=\mathbf{0}$, $\bm{\gamma}_i^0 =\mathbf{0}$; set step size $\rho>0$;
		\STATE Draw $L$ i.i.d. samples $\{\bm{\omega}_l\}_{l=1}^L$ from $p_\kappa(\bm{\omega})$ according to a common random seed.
		\STATE Construct $\{\bm{\phi}_L(\mathbf{x}_{i,t})\}_{t=1}^{T_i}$ using the random features $\{\bm{\omega}_l\}_{l=1}^L$ via \eqref{eq:realRF1} or \eqref{eq:realRF2}.
		\FOR {iterations $k = 1, 2,\cdots$}
		\STATE Update local variable $\bm{\theta}_i^k$ by \eqref{eq:ADMM_update_a};  
		\STATE Transmit $\bm{\theta}_i^k$ to all neighbor $n\; (n\in \mathcal{N}_i)$ and receive $\bm{\theta}_n^k$ from all neighbor $n$;
		\STATE Update local dual variable $\bm{\gamma}_i^k$ by \eqref{eq:ADMM_update_b}.
		\ENDFOR
	\end{algorithmic}
\end{algorithm}

\subsection{COKE: Communication-censored decentralized kernel learning}
\label{subsec:COKE}
From Sections \ref{subsec:RF_kernel} and \ref{subsec:RF_Dectr}, we can see that decentralized kernel learning in the RF space under the consensus optimization framework has much reduced computational complexity, thanks to the RF mapping technique that transforms the learning model into a smaller RF space. In this subsection, we consider the case when the communication resource is limited and we aim to further reduce the communication cost of DKLA. To start, we notice that in Algorithm~\ref{tab:dis_RF_ADMM}, each agent $i\; (i\in \mathcal{N})$ maintains $2+|\mathcal{N}_i|$ local variables at iteration $k$, i.e., its local primal variable $\bm{\theta}_i^k$, local dual variable $\bm{\gamma}_i^k$ and $|\mathcal{N}_i|$ state variables $\bm{\theta}_n^k$ received from its neighbors. While the dual variable $\bm{\gamma}_i^k$ is kept locally for agent $i$, the transmission of its updated local variable $\bm{\theta}_i^k$ to its one-hop neighbors happens in every iteration, which consumes a large amount of communication bandwidth and energy along iterations for large-scale networks. In order to improve the communication efficiency, we develop the COKE algorithm by employing a censoring function at each agent to decide if a local update is informative enough to be transmitted.

To evaluate the importance of a local update at iteration $k$ for agent $i\; (i\in \mathcal{N})$, we introduce a new state variable $\hat{\bm{\theta}}_i^{k-1}$ to record agent $i$'s latest broadcast primal variable up to time $k-1$. Then, at iteration $k$, we define the difference between agent $i$'s current state $\bm{\theta}_i^k$ and its previously transmitted state $\hat{\bm{\theta}}_i^{k-1}$ as
\begin{equation}
\label{eq:difference}
\bm{\xi}_i^k = \hat{\bm{\theta}}_i^{k-1}-\bm{\theta}_i^k,
\end{equation}
and choose a censoring function as 
\begin{equation}
\label{eq:cens_func}
H_i(k,\bm{\xi}_i^k)=\|\bm{\xi}_i^k\|_2-h_i(k),
\end{equation}
where $\{h_i(k)\}$ is a non-increasing non-negative sequence. A typical choice for the censoring function is
$H_i(k,\bm{\xi}_i^k)=\|\bm{\xi}_i^k\|_2 - v\mu^k,$ where $\mu\in(0,1)$ and $v>0$ are constants. When $H_i(k,\bm{\xi}_i^k) <0$, $\bm{\theta}_i^k$ is deemed not informative enough, and hence will not be transmitted to its neighbors.  

When executing the COKE algorithm, each agent $i$ maintains $3+|\mathcal{N}_i|$ local variables at each iteration $k$. Comparing with the DKLA update in~\eqref{eq:ADMM_update}, the additional local variable is the state variable $\hat{\bm{\theta}}_i^k$ that records its latest broadcast primal variable up to time $k$. Moreover, the $|\mathcal{N}_i|$ state variables from its neighbors are $\hat{\bm{\theta}}_n^k$ that record the latest received primal variables from its neighbors, instead of the timely updated and broadcast variables $\bm{\theta}_n^k$ of its neighbors $n\in\mathcal{N}_i$. While in COKE, each agent computes local updates at every step, its transmission to neighbors does not always occur, but is determined by the censoring criterion \eqref{eq:cens_func}. To be specific, at each iteration $k$, if $H_i(k,\bm{\xi}_i^k)\geq 0$, then $\hat{\bm{\theta}}_i^k=\bm{\theta}_i^k$, and agent $i$ is allowed to transmit its local primal variable $\bm{\theta}_i^k$ to its neighbors. Otherwise, $\hat{\bm{\theta}}_i^k=\hat{\bm{\theta}}_i^{k-1}$ and no information is transmitted. If agent $i$ receives $\bm{\theta}_n^k$ from any neighbor $n$, then that neighbor's state variable kept by agent $i$ becomes $\hat{\bm{\theta}}_n^k=\bm{\theta}_n^k$, otherwise, $\hat{\bm{\theta}}_n^k=\hat{\bm{\theta}}_n^{k-1}$. Consequently, agent $i$'s local parameters are updated as follows:
\begin{subequations} 
	\label{eq:update_COKE}
	\begin{align} 
	&\bm{\theta}_i^k := \arg \min_{\bm{\theta}_i} \;\left \{\hat{R}_i(\bm{\theta}_i)+\rho|\mathcal{N}_i| \|\bm{\theta}_i\|_2^2 + \bm{\theta}_i^\top\left[\bm{\gamma}_i^{k-1}-\rho\sum_{n\in\mathcal{N}_i}\left(\hat{\bm{\theta}}_i^{k-1}+ \hat{\bm{\theta}}_n^{k-1}\right)\right]\right\},  	\label{eq:update_COKE_a}\\
	&\bm{\gamma}_i^k =\bm{\gamma}_i^{k-1}+\rho\sum_{n\in\mathcal{N}_i}\Big(\hat{\bm{\theta}}_i^k- \hat{\bm{\theta}}_n^k\Big)	\label{eq:update_COKE_b},
	\end{align} 
\end{subequations}
with a censoring step conducted between \eqref{eq:update_COKE_a} and \eqref{eq:update_COKE_b}. 
We outline the COKE algorithm in Algorithm \ref{tab:COKE}.

\begin{algorithm}[t]
	\caption{COKE Run at Agent $i$ }
	\label{tab:COKE}
	\begin{algorithmic}[1]
		\REQUIRE Kernel $\kappa$, the number of random features $L$, the censoring thresholds $\{h_i(k)\}$, and $\lambda$ to control over-fitting; initialize local variables to $\bm{\theta}_i^0=\mathbf{0}$, $\hat{\bm{\theta}}_i^0=\mathbf{0}$, $\bm{\gamma}_i^0 =\mathbf{0}$; set step size $\rho>0$;
		\STATE Draw $L$ i.i.d. samples $\{\bm{\omega}_l\}_{l=1}^L$ from $p_\kappa(\bm{\omega})$ according to a common random seed.
		\STATE Construct $\{\bm{\phi}_L(\mathbf{x}_{i,t})\}_{t=1}^{T_i}$ using the random features $\{\bm{\omega}_l\}_{l=1}^L$ via \eqref{eq:realRF1} or \eqref{eq:realRF2}.
	 
		\FOR {iterations $k = 1, 2,\cdots$} 
	 
		\STATE Update local variable $\bm{\theta}_i^k$ by \eqref{eq:update_COKE_a};  
		\STATE Compute $\bm{\xi}_i^k = \hat{\bm{\theta}}_i^{k-1}-\bm{\theta}_i^k$;
		\STATE If $H_i(k,\bm{\xi}_i^k) = \|\bm{\xi}_i^k\|_2-h_i(k) \geq 0$, transmit $\bm{\theta}_i^k$ to neighbors and let $\hat{\bm{\theta}}_i^k=\bm{\theta}_i^k$; else do not transmit and let $\hat{\bm{\theta}}_i^k=\hat{\bm{\theta}}_i^{k-1}$;
		\STATE If receives $\bm{\theta}_n^k$ from neighbor $n$, let $\hat{\bm{\theta}}_n^k=\bm{\theta}_n^k$; else let $\hat{\bm{\theta}}_n^k=\hat{\bm{\theta}}_n^{k-1}$;
		\STATE Update local dual variable $\bm{\gamma}_i^k$ by \eqref{eq:update_COKE_b}.
 
		\ENDFOR
	\end{algorithmic}
\end{algorithm}

The key feature of COKE is that agent $i$'s local variables $\bm{\theta}_i^k$ and $\bm{\gamma}_i^k$ are updated all the time, but the transmission of $\bm{\theta}_i^k$ occurs only when the censoring condition is met. By skipping unnecessary transmissions, the communication efficiency of COKE is improved. It is obvious that large $\{h_i(k)\}$ saves more communication but may lead to divergence from the optimal solution $\bm{\theta}^\ast$ of \eqref{eq:empirical_risk_theta_N}, while small $\{h_i(k)\}$ does not contribute much to communication saving. Noticeably, DKLA is a special case of COKE when the communication censoring strategy is absent by setting $h_i(k) = 0, \forall i,k$.

\section{Theoretical Guarantees}
\label{sec:analytic}
In this section, we perform theoretical analyses to address two questions related to the convergence properties of DKLA and COKE algorithms. First, do they converge to the globally optimal point, and if so, at what rate? Second, what is their achieved generalization performance in learning? Since DKLA is a special case of COKE, the analytic results of COKE, especially the second one, extend to DKLA straightforwardly. For theoretical analysis, we make the following assumptions.

\begin{assum}
	\label{ass:net_connect}
	The network with topology $\mathcal{G}=(\mathcal{N}, \mathcal{C}, \mathbf{A})$ is undirected and connected.
\end{assum}

\begin{assum} 
	\label{ass:strong_convex}
	The local cost functions $\hat{R}_i$ are strongly convex with constants $m_{\hat{R}_i}>0$ such that $\forall i\in\mathcal{N}$, $ \langle \nabla \hat{R}_i(\tilde{\bm{\theta}}_a)- \nabla \hat{R}_i(\tilde{\bm{\theta}}_b), \tilde{\bm{\theta}}_a-\tilde{\bm{\theta}}_b\rangle\geq m_{\hat{R}_i}\|\tilde{\bm{\theta}}_a-\tilde{\bm{\theta}}_b\|_2^2$, for any $\tilde{\bm{\theta}}_a, \tilde{\bm{\theta}}_b \in \mathbb{R}^{L}$. The minimum convexity constant is $m_{\hat{R}}:=\min_{i} m_{\hat{R}_i}$. The gradients of the local cost functions are Lipschitz continuous with constants $M_{\hat{R}_i}>0, \forall\; i$. That is, $\|\nabla \hat{R}_i(\tilde{\bm{\theta}}_a)- \nabla \hat{R}_i(\tilde{\bm{\theta}}_b)\|_2\leq M_{\hat{R}_i}\|\tilde{\bm{\theta}}_a-\tilde{\bm{\theta}}_b\|_2$ for any agent $i$ given any $\tilde{\bm{\theta}}_a, \tilde{\bm{\theta}}_b \in \mathbb{R}^{L}$. The maximum Lipschitz constant is $M_{\hat{R}}:=\max_{i} M_{\hat{R}_i}$.
\end{assum}

\begin{assum}
	\label{ass:sample_differ_little}
	The number of training samples of different agents is of the same order of magnitude, i.e., $\frac{\max_{i} T_i - \min_{i} T_i}{\min_{i}T_i}<10, \forall i\in \mathcal{N}$.
\end{assum}

\begin{assum}
	\label{ass:f_H-exist}
	There exists $f_\mathcal{H}\in \mathcal{H}$, such that for all estimators $f\in\mathcal{H}$, $\mathcal{E}(f_\mathcal{H})\leq \mathcal{E}(f)$, where $\mathcal{E}(f):=\mathbb{E}_p\left[\ell(f(\mathbf{x}),y) \right]$ is the expected risk to measure the generalization ability of the estimator $f$. 
\end{assum} 
 
Assumption~\ref{ass:net_connect} and \ref{ass:strong_convex} are standard for decentralized optimization over decentralized networks~\citep{shi2014linear}, Assumption~\ref{ass:f_H-exist} is standard in generalization performance analysis of kernel learning~\citep{li2018towards}, and Assumption~\ref{ass:sample_differ_little} is enforced to exclude the case of extremely unbalanced data distributed over the network.

\subsection{Linear convergence of DKLA and COKE}   

We first establish that DKLA enables agents in the decentralized network to reach consensus on the prediction function at a linear rate. We then show that when the censoring function is properly chosen and the penalty parameter satisfies certain conditions, COKE also guarantees that the individually learned functional on the same sample linearly converges to the optimal solution.

\noindent
{\bf Theorem 1} {\it
	\textbf{[Linear convergence of DKLA]} Initialize the dual variables as $\bm{\gamma}_i^0=\mathbf{0},\;\forall i$, with Assumptions~\ref{ass:net_connect} - \ref{ass:sample_differ_little}, the learned functional at each agent through DKLA is R-linearly convergent to the optimal functional $\hat{f}_{\bm{\theta}^\ast} (\mathbf{x}):= (\bm{\theta}^\ast)^\top \bm{\phi}_L(\mathbf{x})$ for any $\mathbf{x}\in\mathcal{X}$, where $\bm{\theta}^\ast$ denotes the optimal solution to~\eqref{eq:empirical_risk_theta_N} obtained in the centralized case. That is,  
	\begin{equation}
	\label{eq:consensus_dkla}
	\lim_{k\rightarrow\infty}  \hat{f}_{\bm{\theta}_i^k}(\mathbf{x}) = \hat{f}_{\bm{\theta}^\ast} (\mathbf{x}), \forall i.
	\end{equation} 
	
\noindent
{\bf Proof}.} See Appendix A.

\noindent
{\bf Theorem 2} {\it
	\textbf{[Linear convergence of COKE]} 	
	Initialize the dual variables as $\bm{\gamma}_i^0=\mathbf{0},\;\forall i$, set the censoring thresholds to be $h(k) = v\mu^k$, with $v>0$ and $\mu\in (0,1)$, and choose the penalty parameter $\rho$ such that 
	\begin{equation}
	\label{eq：choose_rho}
	\begin{split}
	0&<\rho  <\min \left\{ \frac{4m_{\hat{R}}}{\eta_1},\frac{(\nu-1)\tilde{\sigma}_{\min}^2 (\mathbf{S}_{-})}{\nu \eta_3 \tilde{\sigma}_{\max}^2(\mathbf{S}_{+})}, \left(\frac{\eta_1}{4}+\frac{\eta_2 \tilde{\sigma}_{\max}^2(\mathbf{S}_{+})}{8}\right)^{-1}\left(m_{\hat{R}}-\frac{\eta_3\nu M_{\hat{R}}^2 }{\tilde{\sigma}_{\min}^2(\mathbf{S}_{-})}    \right) \right\},
	\end{split}
	\end{equation}
	where $\eta_1>0$, $\eta_2>0$, $\eta_3>0$ and $\nu>1$ are arbitrary constants, $m_{\hat{R}}$ and $M_{\hat{R}}$ are the minimum strong convexity constant of the local cost functions and the maximum Lipschitz constant of the local gradients, respectively. $\tilde{\sigma}_{\max}(\mathbf{S}_{+})$ and $\tilde{\sigma}_{\min}(\mathbf{S}_{-})$ are the maximum singular value of the unsigned incidence matrix $\mathbf{S}_{+}$ and the minimum non-zero singular value of
	the signed incidence matrix $\mathbf{S}_{-}$ of the network, respectively. Then, with Assumptions~\ref{ass:net_connect} - \ref{ass:sample_differ_little}, the learned functional at each agent through COKE is R-linearly convergent to the optimal one $\hat{f}_{\bm{\theta}^\ast} (\mathbf{x}):=(\bm{\theta}^\ast)^\top \bm{\phi}_L(\mathbf{x})$ for any $\mathbf{x}\in\mathcal{X}$, where $\bm{\theta}^\ast$ denotes the optimal solution to~\eqref{eq:empirical_risk_theta_N} obtained in the centralized case. That is, 
	\begin{equation}
	\label{eq:consensus_f}
	\lim_{k\rightarrow\infty}  \hat{f}_{\bm{\theta}_i^k}(\mathbf{x})	 = \hat{f}_{\bm{\theta}^\ast} (\mathbf{x}), \forall i.
	\end{equation} } \hfill\BlackBox

\noindent
{\bf Proof}. See Appendix A.

\noindent
\textbf{Remark 1.} It should be noted that the kernel transformation with RF mapping is essential in enabling convex consensus formulation with convergence guarantee. For example, in a regular optimization problem with a local cost function $(y-f(\mathbf{x}))^2$, even if it is quadratic, the nonlinear function $f(\mathbf{x})$ inside destroys the convexity. In contrast, with RF mapping, $f(\mathbf{x})$ of any form is expressed as a linear function of $\bm{\theta}$, and hence the local cost function is guaranteed to be convex. For decentralized kernel learning, many widely-adopted loss functions result in (strongly) convex local objective functions in the RF space, such as the quadratic loss in a regression problem and logistic loss in a classification problem.

\noindent
\textbf{Remark 2.} For Theorem 2, notice that choosing larger $v$ and $\mu$ in the design of the censoring thresholds in COKE leads to less communication per iteration at the expense of possible performance degradation, whereas smaller $v$ and $\mu$ may not contribute much to communication saving. However, it is challenging to acquire an explicit tradeoff between communication cost and steady-state accuracy, since the designed censoring thresholds do not have an explicit relationship with the update of the model parameter.

The above theorems establish the exact convergence of the functional learned in the multi-agent system for the decentralized kernel regression problem via DKLA and COKE. Different from previous works~\citep{koppel2018decentralized, shin2018distributed}, our analytic results are obtained by converting the non-parametric data-dependent learning model into a parametric data-independent model in the RF space and solved under the consensus optimization framework. In this way, we not only reduce the computational complexity of the standard kernel methods and make the RF-based kernel methods scalable to large-size datasets, but also protect data privacy since no raw data exchange among agents is required and the RF mapping is not one-to-one mapping. RF mapping is crucial in our algorithms, with which we are able to show the linear convergence of the functional by showing the linear convergence of the iteratively updated decision variables in the RF space; see Appendix A for more details.   
 
\subsection{Generalization property of COKE}
The ultimate goal of decentralized learning is to find a function that generalizes well for the ensemble of all data from all agents. To evaluate the generalization property of the predictive function learned by COKE, we are then interested in bounding the difference between the expected risk of the predictive function learned by COKE at the $k$-th iteration, defined as $\mathcal{E}(\hat{f}^k):=\sum_{i=1}^N \mathcal{E}_i(\hat{f}_{\bm{\theta}_i^k}):=\sum_{i=1}^N \mathbb{E}_p [(y-(\bm{\theta}_i^k)^\top \bm{\phi}_L(\mathbf{x}))^2] 
$, and the expected risk $\mathcal{E}(f_{\mathcal{H}})$ in the RKHS. This is different from bounding the approximation error between the kernel $\kappa$ and the approximated $\hat{\kappa}_L$ by $L$ random features as in the literature~\citep{rahimi2008random, sutherland2015error, sriperumbudur2015optimal}. As DKLA is a special case of COKE, the generalization performance of COKE can be extended to DKLA straightforwardly.

To illustrate our finding, we focus on the kernel regression problem whose loss function is least squares, i.e., $\ell(y,f(\mathbf{x}))=(y-f(\mathbf{x}))^2$. With RF mapping, the objective function \eqref{eq:empirical_risk_theta_N} of the regression problem can be formulated as
\begin{equation}
\label{eq:obj_RF}
\hat{R}(\bm{\theta}) = \sum_{i=1}^N \hat{R}_i(\bm{\theta})=\sum_{i=1}^N \left(\frac{1}{T_i}\|\mathbf{y}_i-(\bm{\Phi}_{L}^i)^\top\bm{\theta}\|_2^2+\frac{\lambda}{N}  \|\bm{\theta}\|_2^2\right),
\end{equation} 
where $\mathbf{y}_i = [y_{i,1},\dots,y_{i,T_i}]^\top \in \mathbb{R}^{T_i\times 1}$,  $\bm{\Phi}_{L}^i = [\bm{\phi}_{L}(\mathbf{x}_{i,1}),\dots, \bm{\phi}_{L}(\mathbf{x}_{i,T_i})]\in\mathbb{R}^{L \times T_i}$, and $\bm{\phi}_{L}(\mathbf{x}_{i,t})$ is the data mapped to the RF space.

The optimal solution of \eqref{eq:obj_RF} is given in closed form by 
\begin{equation}
\label{eq:optimal_KRR_RF}
\bm{\theta}^\ast = (\tilde{\bm{\Phi}}^\top \tilde{\bm{\Phi}} +\lambda \mathbf{I})^{-1} \tilde{\bm{\Phi}}^\top \tilde{\mathbf{y}},
\end{equation}
where $\tilde{\bm{\Phi}}=[\tilde{\bm{\Phi}}_L^1,\dots, \tilde{\bm{\Phi}}_L^N]^\top\in \mathbb{R}^{T\times L}$ with $\tilde{\bm{\Phi}}_L^i = \frac{1}{\sqrt{T_i}}\bm{\Phi}_{L}^i, \forall i\in\mathcal{N}$,  and $\tilde{\mathbf{y}}=[\tilde{\mathbf{y}}_1; \dots;\tilde{\mathbf{y}}_N]\in \mathbb{R}^{T \times 1}$ with $\tilde{\mathbf{y}}_i = \frac{1}{\sqrt{T_i}}\mathbf{y}_i, \forall i\in\mathcal{N}$. The optimal prediction model is then expressed by 
\begin{equation}
\label{eq:optimal_KRR_f}
\hat{f}_{\bm{\theta}^\ast} (\mathbf{x})=(\bm{\theta}^\ast)^\top \bm{\phi}_L(\mathbf{x}).
\end{equation}

In the following theorem, we give a general result of the generalization performance of the predictive function learned by COKE for the kernel regression problem, which is built on the linear convergence result given in Theorem 3 and taking into account of the number of random features adopted.

\noindent
{\bf Theorem 3} {\it
	Let $\lambda_\mathbf{K}$ be the largest eigenvalue of the kernel matrix $\mathbf{K}$ constructed by all data, $\mathbf{X}_T$, and choose the regularization parameter $\lambda<\lambda_\mathbf{K}/T$ so as to control overfitting. Under the Assumptions \ref{ass:net_connect} - \ref{ass:f_H-exist}, with the censoring function and other parameters given in Theorem 2, for all $\delta_p \in (0,1)$ and $\|f\|_\mathcal{H}\leq 1$, if the number of random features $L$ satisfies 
	\begin{equation}
	\label{eq:RF_number_KRR}
	L \geq  \frac{1}{\lambda}(\frac{1}{\epsilon^2}+\frac{2}{3\epsilon})\log \frac{16d_{\mathbf{K}}^\lambda}{\delta_p}\nonumber,
	\end{equation} 
	then with probability at least $1-\delta_p$, the excess risk of $\mathcal{E} (\hat{f}^k) $ obtained by Algorithm \ref{tab:COKE} converges to an upper bound, i.e., 
	\begin{equation}
	\lim_{k\rightarrow\infty} (\mathcal{E} (\hat{f}^k) - \mathcal{E}(f_\mathcal{H})) \leq 3\lambda + O(\frac{1}{\sqrt{T}}),
	\end{equation}
	where $\epsilon \in (0,1)$, and $d_{\mathbf{K}}^\lambda:=\mathrm{Tr}(\mathbf{K}(\mathbf{K}+\lambda T \mathbf{I})^{-1})$ is the number of effective degrees of freedom that is known to be an indicator of the number of independent parameters in a learning problem~\citep{avron2017random}.} \hfill\BlackBox

\noindent
{\bf Proof}. See Appendix B.

Theorem 3 states the tradeoff between the computational efficiency and the statistical efficiency through the regularization parameter $\lambda$, effective dimension $d_{\mathbf{K}}^\lambda$, and the number of random features adopted. We can see that to bound the excess risk with a higher probability, we need more random features, which results in a higher computational complexity. The regularization parameter is usually determined by the number of training data and one common practice is to set $\lambda = O(1/\sqrt{T})$ for the regression problem~\citep{caponnetto2007optimal}. Therefore, with $O(\sqrt{T}\log d_{\mathbf{K}}^\lambda)$ features, COKE achieves a learning risk  of $O(1/\sqrt{T})$ at a linear rate. We also notice that different sampling strategies affect the number of random features required to achieve a given generalization error. For example, importance sampling is studied for the centralized kernel learning in RF space in~\citep{li2018towards}. Interested readers are referred to~\citep{li2018towards} and references therein.

\section{Experiments}
\label{sec:exper}
This section evaluates the performance of our COKE algorithm in regression tasks using both synthetic and real-world datasets. Since we consider the case that data are only locally available and cannot be shared among agents, the following RF-based methods are used to benchmark our COKE algorithm. 

\noindent
\textbf{CTA.} This is a form of diffusion-based technique where all agents first construct their RF-mapped data $\{\bm{\phi}_L(\mathbf{x}_{i,t})\}_{t=1}^{T_i}$, for $t =1, \dots, T_i, \forall i$, using the same random features $\{\bm{\omega}_l\}_{l=1}^L$ as DKLA and COKE. Then at each iteration $k$, each agent $i$ first combines information from its neighbors, i.e., $\bm{\theta_n}, \forall n\in\mathcal{N}_i$ with its own parameter $\bm{\theta}_i$ by aggregation. Then, it updates its own parameter $\bm{\theta}_i$ using the gradient descent method with the aggregated information~\citep{sayed2014adaptation}. The cost function for agent $i$ is given in \eqref{eq:empirical_risk_theta_i}. Note that this method has not been formally proposed in existing works for RF-based decentralized kernel learning with batch-form data, but we introduced it here only for comparison purpose. An online version that deals with streaming data is available in~\citep{bouboulis2018online}. The batch version of CTA introduced here is expected to converge faster than the online version.  
 
\noindent
\textbf{DKLA.} Algorithm \ref{tab:dis_RF_ADMM} proposed in Section \ref{subsec:RF_Dectr} where ADMM is applied and the communication among agents happen at every iteration without being censored.

The performance of all algorithms is evaluated using both synthetic and real-world datasets, where the entries of data samples are normalized to lie in $[0,1]$ and each agent uses $70\%$ of its data for training and the rest for testing. The learning performance at each iteration is evaluated using mean-squared-error (MSE) given by $\mathrm{MSE}(k) =\frac{1}{T}\sum_{i=1}^N\sum_{t=1}^{T_i}(y_{i,t}-(\bm{\theta}_i^k)^\top\bm{\phi}_L(\mathbf{x}_{i,t}))^2$. The decision variable $\boldsymbol{\theta}_i$ for CTA is initialized as $\boldsymbol{\theta}_i^0 = \mathbf{0}, \forall i$ as that in DKLA and COKE. For COKE, it should be noted that the design of the censoring function is crucial. For the censoring thresholds adopted in Theorem 2, choosing larger $v$ and $\mu$ to design the censor thresholds leads to less communication per iteration but may result in performance degradation. For all simulations, the kernel bandwidth is fine-tuned for each dataset individually via cross-validation. The parameters of the censoring function are tuned to achieve the best learning performance at nearly no performance loss.  

\subsection{Synthetic dataset}
In this setup, the connected graph is randomly generated with $N=20$ nodes and $95$ edges. The probability of attachment per node equals to $0.3$, i.e., any pair of two nodes are connected with a probability of 0.3. 
Each agent has~$T_i\in(4000,6000)$ data pairs generated following the model~$y_{i,t} = \sum_{m=1}^{50} b_m\kappa(\mathbf{c}_m,\mathbf{x}_{i,t}) + e_{i,t}$, where~$b_m$ are uniformly drawn from $[0,1]$, $\mathbf{c}_m \!\sim \!\mathbf{N}(\bm{0},\mathbf{I}_{5})$, $\mathbf{x}_{i,t} \sim \mathbf{N}(\bm{0}, \mathbf{I}_{5})$, and $e_{i,t} \sim \mathbf{N}(0, 0.1)$. The kernel $\kappa$ in the model is Gaussian with a bandwidth $\sigma =  5$. 

\subsection{Real datasets}
To further evaluate our algorithms, the following popular real-world datasets from UCI machine learning repository are chosen~\citep{asuncion2007uci}.

\noindent
\textbf{Tom's hardware.} This dataset contains $T=11000$ samples with $\mathbf{x}_t\in \mathbb{R}^{96}$ whose features include the number of created discussions and authors interacting on a topic and $y_t\in \mathbb{R}$ representing the average number of displays to a visitor about that topic~\citep{kawala2013predictions}.  

\noindent
\textbf{Twitter.} This dataset consists of $T=13800$ samples with $\mathbf{x}_t\in \mathbb{R}^{77}$ being a feature vector reflecting the number of new interactive authors and the length of discussions on a given topic, etc., and $y_t\in \mathbb{R}$ representing the average number of active discussion on a certain topic. The learning task is to predict the popularity of these topics. We also include a larger Twitter dataset for testing which has $T=98704$ samples~\citep{kawala2013predictions}. 

\noindent
\textbf{Energy.} This dataset contains $T = 19735 $ samples with $\mathbf{x}_t\in \mathbb{R}^{28}$ describing the
humidity and temperature in different areas of the houses, pressure, wind speed and viability outside, while $y_t$ denotes the total energy consumption in the house~\citep{candanedo2017data}. 

\noindent
\textbf{Air quality.} This dataset contains dataset collects $T = 9358$ samples measured by a gas multi-sensor device in an Italian city, where $\mathbf{x}_t\in \mathbb{R}^{13}$ represents the hourly concentration of CO, NOx, NO2, etc, while $y_t$ denotes the concentration of polluting chemicals in the air~\citep{de2008field}.

\subsection{Parameter setting and performance analysis}
For synthetic data, we adopt a Gaussian kernel with a bandwidth $\sigma =  1$ for training and use $L = 100$ random features for kernel approximation. Note that the chosen $\sigma$ differs from that of the actual data model. The censoring thresholds are $h(k)=0.95^k$, the regularization parameter $\lambda$ and stepsize $\rho$ of DKLA and COKE are set to be $5\times10^{-5}$ and $10^{-2}$, respectively. The stepsize of CTA is set to be $\eta = 0.99$, which is tuned to achieve the same level of learning performance as COKE and DKLA at its fastest speed.

To show the performance of all algorithms on real datasets concisely and comprehensively, we present the experimental results on the Twitter dataset with $T=13800$ samples by figures and record the experimental results on the remaining datasets by tables. For the Twitter dataset with $T=13800$ samples, we randomly split it into $10$ mini-batches each with $T_i\in(1200, 1400)$ data pairs while $\sum_{i=1}^{10} T_i = T$. The 10 mini-batches are distributed to 10 agents connected by a random network with $28$ edges. We use $100$ random features to approximate a Gaussian kernel with a bandwidth $\sigma =  1$ during the training process. The parameters $\lambda$ and $\rho$ are set to be $10^{-3}$ and $10^{-2}$, respectively. The censoring thresholds are $h(k)= 0.97^k$. The stepsize of CTA is set to be $\eta = 0.99$ to balance the learning performance and the convergence speed.   

In Fig.1, we show that the individually learned functional at each agent via COKE reaches consensus to the optimal estimate for both synthetic and real datasets. In Fig. 2, we compare the MSE performance of COKE, DKLA, and CTA. Both figures show that COKE converges slower than DKLA due to the communications skipped by the censoring step. However, the learning performance of COKE eventually is the same as DKLA. For the diffusion-based CTA algorithm, it converges the slowest. In Fig. 3, we show the MSE performance versus the communication cost (in terms of the number of transmissions). As CTA converges the slowest and communicates all the time, its communication cost is much higher than that of DKLA, and thus we do not include it in Fig. 3 but rather focus on the communication-saving of COKE over DKLA. We can see that to achieve the same level of learning performance, COKE requires much less communication cost than DKLA. Both the synthetic data and the real dataset show communication saving of around 50\% in Fig. 3 for a given learning accuracy, which corroborate the communication-efficiency of COKE.

\begin{figure}
	\centering
	\subfigure[Synthetic data.]{\label{fig:func_syn}\includegraphics[width=7.55cm]{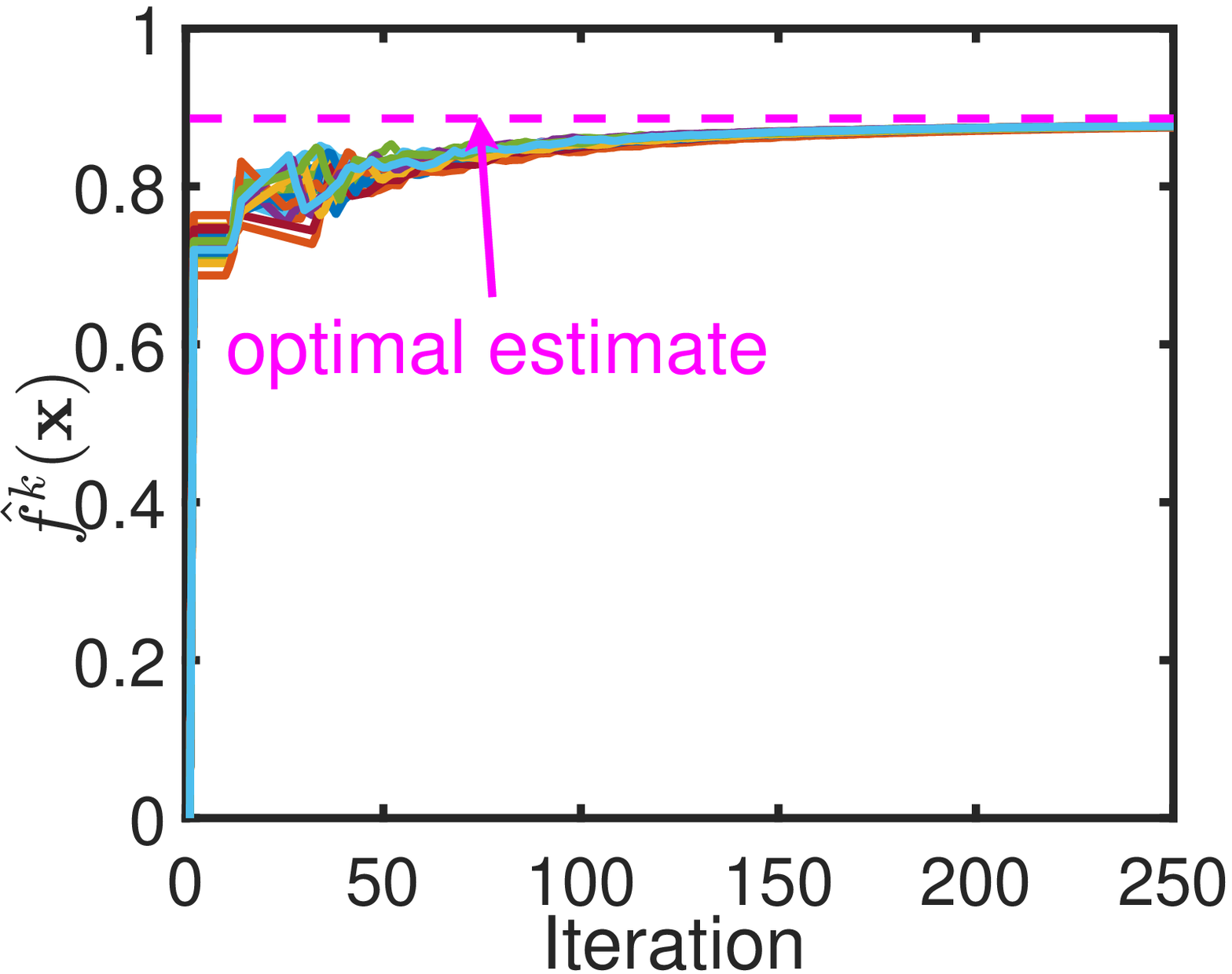}}
	\subfigure[Twitter data.]{\label{fig:func_twitt}\includegraphics[width=7.55cm]{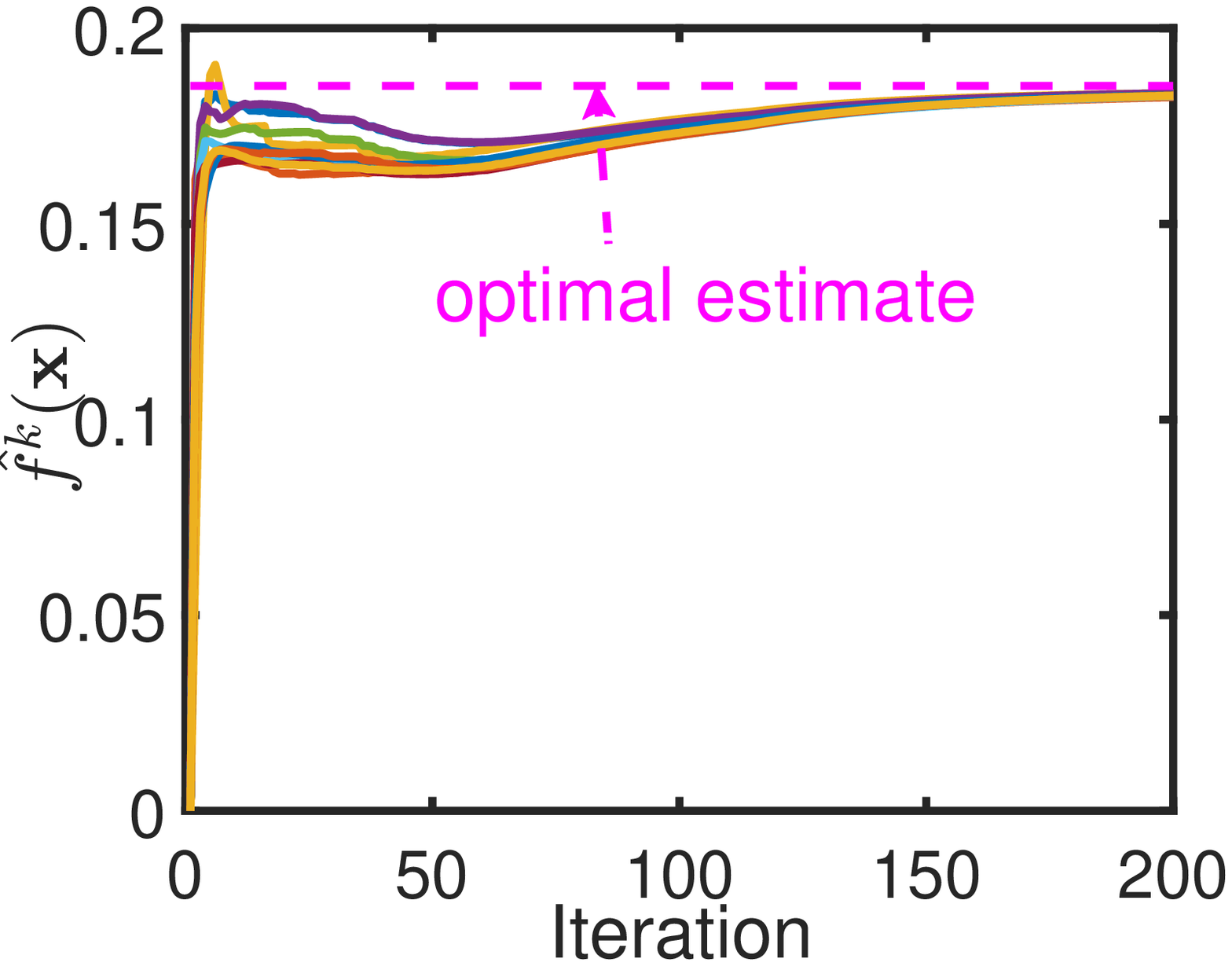}}
	\caption{Functional convergence via COKE for synthetic data (Figure 1 (a)) and the real dataset (Figure 1 (b)). The learned functionals of all distributed agents converge to the optimal estimate where data are assumed to be centrally available.}
\end{figure}
 
\begin{figure}
	\label{fig: MSE}
	\centering
	\subfigure[Synthetic data.]{\label{fig:MSE_iter_syn}\includegraphics[width=7.55cm]{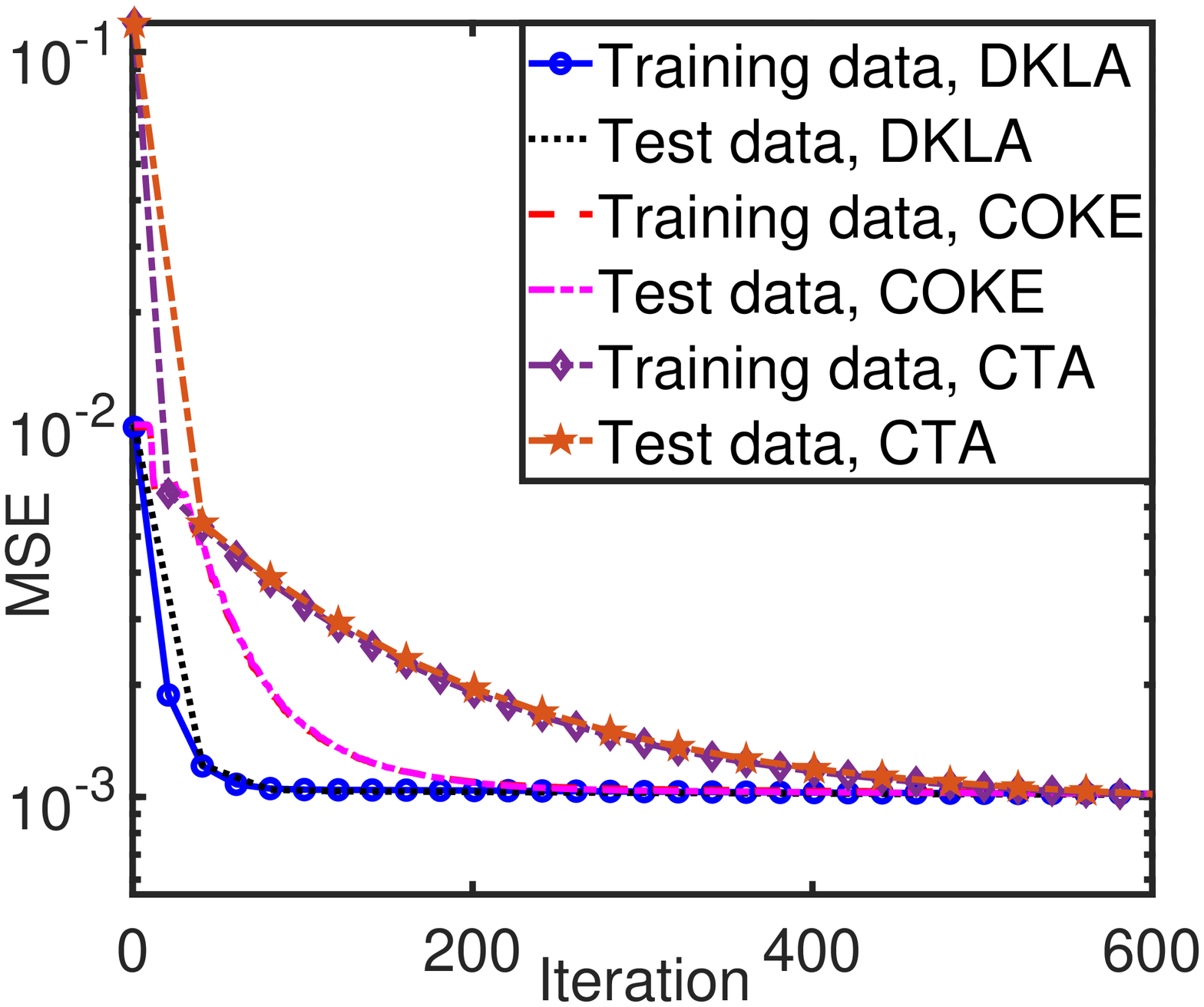}}
	\subfigure[Twitter data.]{\label{fig:MSE_iter_twitt}\includegraphics[width=7.55cm]{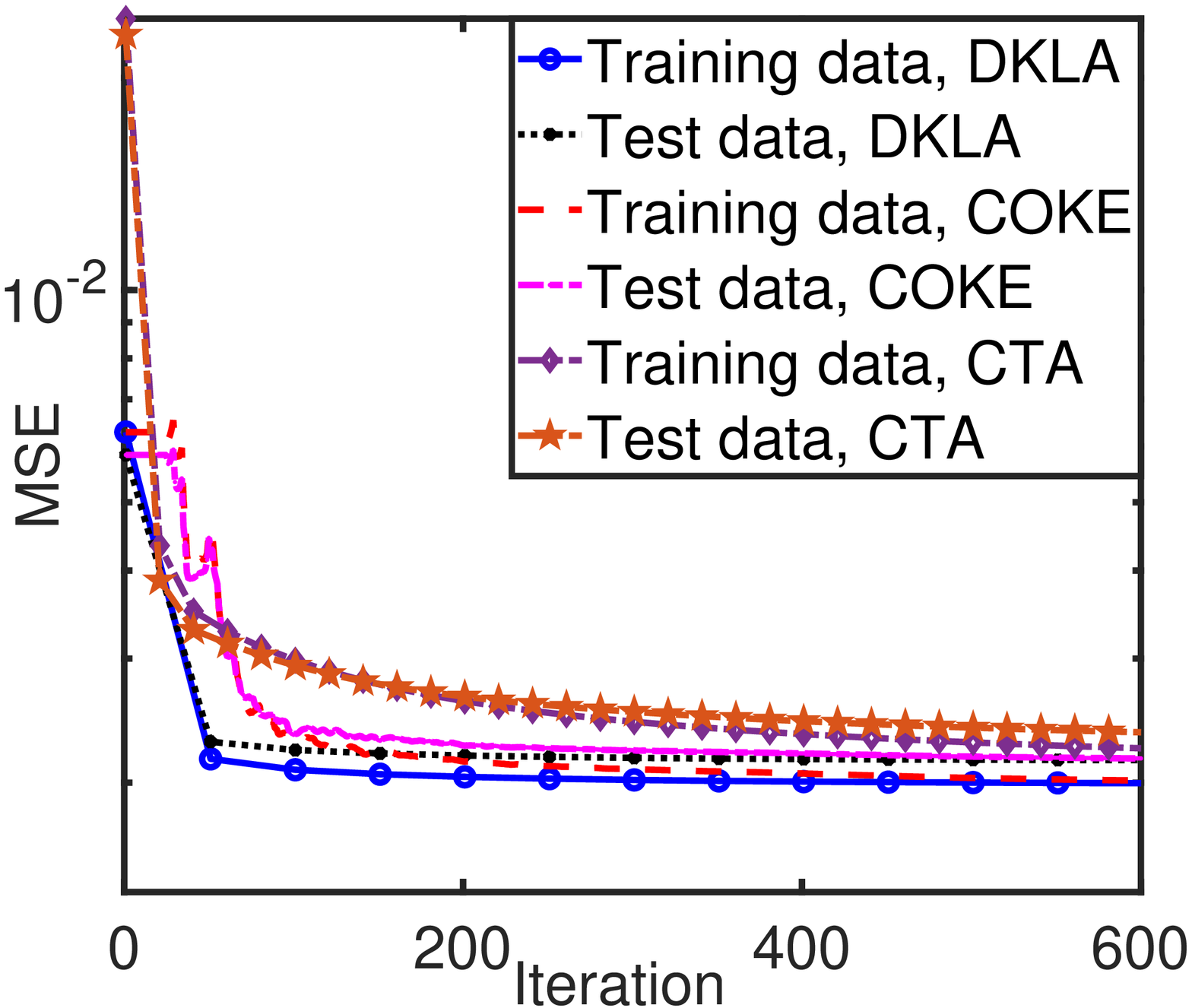}}
	\caption{MSE performance for synthetic data (Figure 2 (a)) and the real dataset (Figure 2 (b)). ADMM-based algorithms (COKE and DKLA) converge faster than the diffusion-based algorithm (CTA) for both synthetic data (Figure 2 (a)) and the real dataset (Figure 2 (b)). Furthermore, DKLA and COKE achieve better learning performance than CTA in terms of MSE on the real dataset.}
\end{figure}

\begin{figure}
	\centering
	\subfigure[Synthetic data.]{\label{fig:MSE_comm_syn}\includegraphics[width=7.55cm]{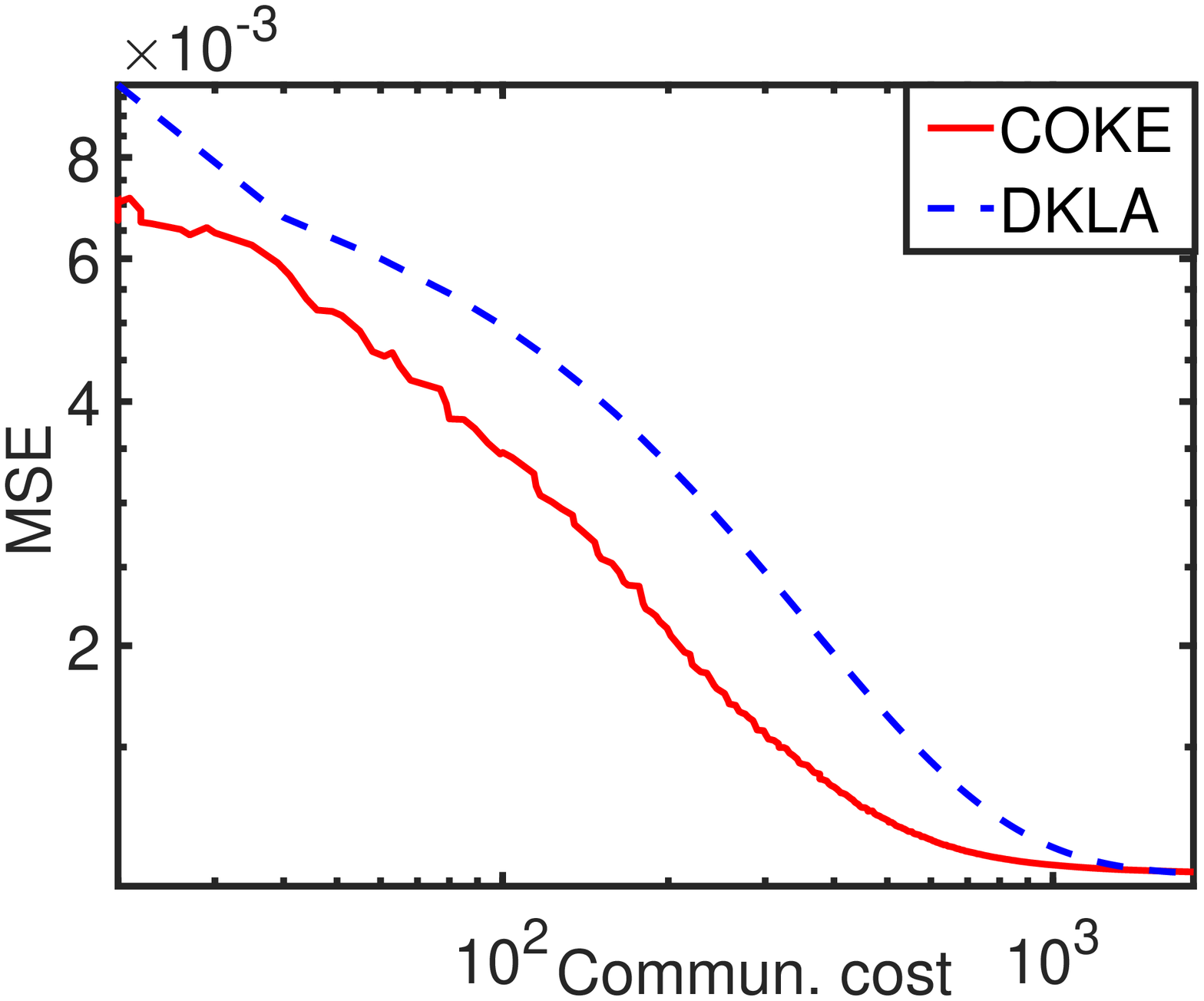}}
	\subfigure[Twitter data.]{\label{fig:MSE_comm_twitt}\includegraphics[width=7.55cm]{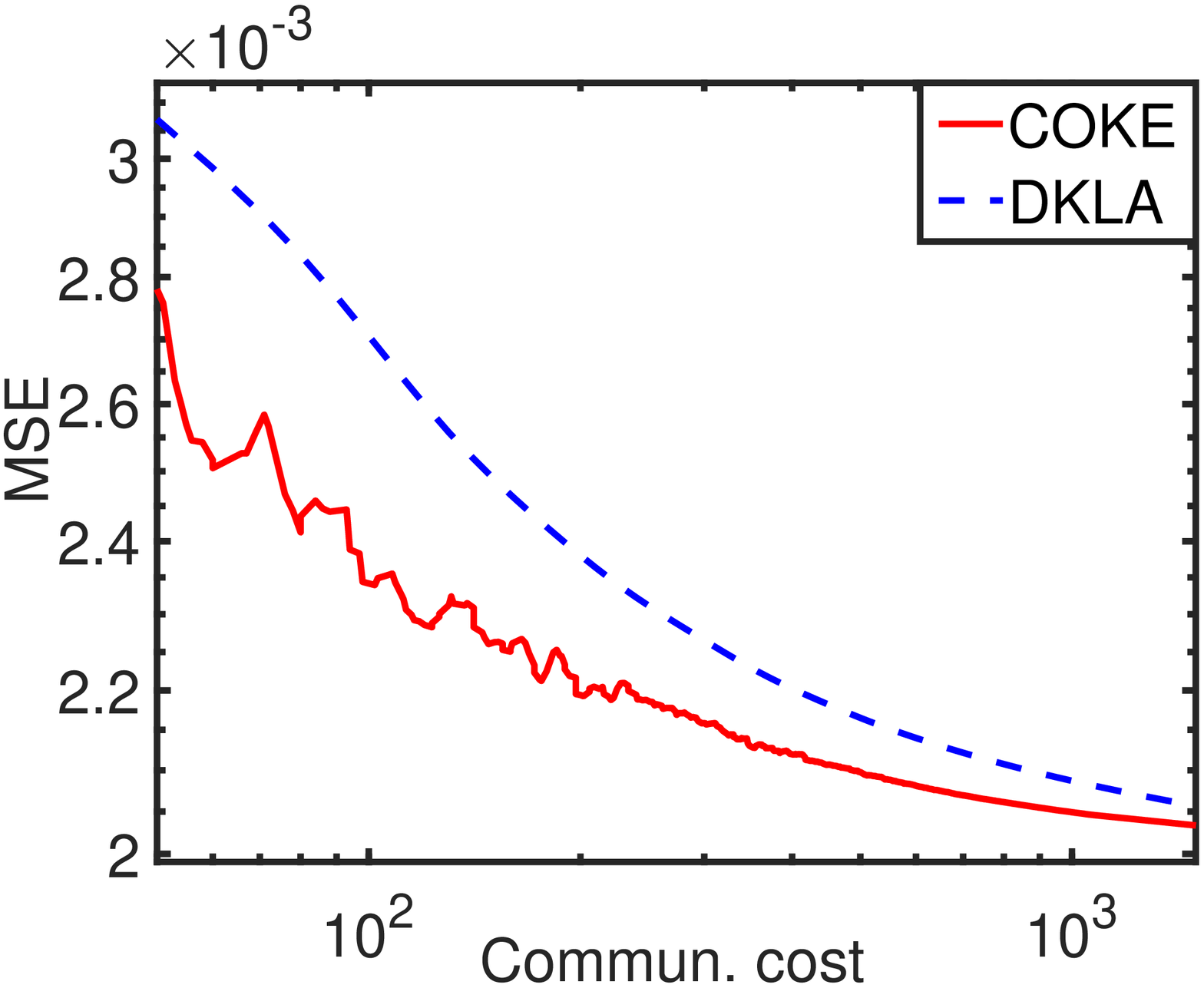}}
	\caption{MSE performance versus communication cost for synthetic data (Figure 3 (a)) and the real dataset (Figure 3 (b)). Compared with DKLA, COKE achieves around 50\% communication saving on the same level of MSE performance for both synthetic data (Figure 3 (a)) and the real dataset (Figure 3 (b)).}
\end{figure}

\begin{table}
	\centering
	\label{tab:TwittL}	
	\begin{tabular}{||c||c|c|c||c|c|c||}
		\hline
		& \multicolumn{3}{|c||}{Training error (MSE ($10^{-3}$)) / Commun. cost} &\multicolumn{3}{|c||}{Test error (MSE($10^{-3}$))} \\
		\hline
		Iteration & CTA   & DKLA   & COKE   & CTA    & DKLA    & COKE  \\
		\hline 	 
		$k=50$   &3.9/500 & 2.4/500 &4.5/\textbf{13} & 4.0 & 2.6 & 4.2  \\
		\hline
		$k=100$ &3.3/1000 & 2.4/1000 & 2.6/\textbf{100} &3.4 &2.6 &2.8 \\
		\hline
		$k=200$ &3.0/2000 & 2.3/2000 & 2.4/\textbf{298} &3.2 &2.5 &2.6\\
		\hline
		$k=500$ &2.7/5000 & 2.3/5000 & \textbf{2.3}/\textbf{902} &2.9 &2.5 &\textbf{2.5}  \\
		\hline
		$k=1000$ &2.5/10000 & 2.2/10000 & \textbf{2.2}/\textbf{4648} &2.7 &2.5 &\textbf{2.5}  \\
		\hline
		$k=1500$ &2.5/15000 & 2.2/15000 & \textbf{2.2}/\textbf{9648} &2.7 &2.5 &\textbf{2.5}  \\
		\hline
		$k=2000$ &2.4/20000 & 2.2/20000 & \textbf{2.2}/\textbf{14648} &2.6 &2.5 &\textbf{2.5}  \\
		\hline
	\end{tabular}
	\caption{MSE performance on the Twitter dataset (large), $\sigma=1$, $L=100$, $\lambda = 10^{-3}$, stepsize $\eta = 0.99$ for CTA, stepsize $\rho=10^{-2}$ for DKLA and COKE, censoring thresholds $h(k) = 0.5\times 0.98^k$. DKLA and COKE achieve better MSE performance than CTA while COKE requires the least communication resource than DKLA.} 	
\end{table}

\begin{table}
	\centering
	\label{tab:Tom}	 	
	\begin{tabular}{||c||c|c|c||c|c|c||}
		\hline
		& \multicolumn{3}{|c||}{Training error (MSE ($10^{-4}$)) / Commun. cost} &\multicolumn{3}{|c||}{Test error (MSE($10^{-4}$))}\\
		\hline
		Iteration & CTA   & DKLA   & COKE   & CTA    & DKLA    & COKE  \\
		\hline 	 
		$k=50$   &20.02/500 & 10.01/500 &17.40/\textbf{10} & 20.16 & 11.20 & 18.82  \\
		\hline
		$k=100$ &16.6/1000 & 9.91/1000 & 10.67/\textbf{112} &17.09 &11.10 &11.86 \\
		\hline
		$k=200$ &13.68/2000 & 9.90/2000 & 9.97/\textbf{331} &14.58 &11.10 &11.15 \\
		\hline
		$k=500$ &11.19/5000 & 9.90/5000 & \textbf{9.90}/\textbf{1114} &12.35 &11.10 &\textbf{11.10}  \\
		\hline
		$k=1000$ &10.27/10000 &  9.90/10000 & \textbf{9.90}/\textbf{5600} &11.47 &11.10 &\textbf{11.10}  \\
		\hline
		$k=1500$ &10.01/15000 &  9.90/15000 & \textbf{9.90}/\textbf{10600} &11.22 &11.10 &\textbf{11.10}  \\
		\hline
		$k=2000$ &9.92/20000 &  9.90/20000 & \textbf{9.90}/\textbf{15600 } &11.13 &11.10 &\textbf{11.10} \\
		\hline
	\end{tabular}
	\caption{MSE performance on the Tom's hardware dataset, $\sigma=1$, $L=100$, $\lambda = 10^{-2}$, stepsize $\eta = 0.99$ for CTA, stepsize $\rho=10^{-2}$ for DKLA and COKE, censoring thresholds $h(k) = 0.5\times 0.95^k$. DKLA and COKE achieve better MSE performance than CTA while COKE requires the least communication resource than DKLA.}	
\end{table} 

The performance of all three algorithms on the rest four datasets is listed in Table 1 - 6. All results show that COKE saves much communication (almost 50\%) within a negligible learning gap from DKLA, and both DKLA and COKE require much less communication resources than CTA. For example, the number of transmissions required to reach a training estimation error of $2.3\times 10^{-3}$ on Twitter dataset by COKE is 577, which is only $53\%$ of that required by DKLA to reach the same level of learning performance. For Tom's hardware dataset, COKE requires 361 total transmissions to reach a learning error of $9.95\times 10^{-4}$, which is $56.4\%$ of DKLA and $0.02\%$ of CTA. Note that much of the censoring occurs at the beginning update iterations. While at the later stage, COKE nearly transmits all parameters at every iteration since the censoring thresholds are smaller than the difference between two consecutive updates.   

\begin{table}
	\centering
	\label{tab:TwittL_CommCost}	
	\begin{tabular}{||c||c|c|c||c||c|c|c||}
		\hline
		\multicolumn{4}{|c||}{Twitter dataset (large)} & \multicolumn{4}{|c||}{Tom's hardware} \\
		\hline 
	    MSE ($10^{-3}$) 	& \multicolumn{3}{|c||}{Commun. cost} & MSE ($10^{-4}$) 	& \multicolumn{3}{|c||}{Commun. cost} \\
		\hline
		& CTA    & DKLA    & COKE &	 & CTA    & DKLA    & COKE   \\
		\hline 	 
		5   & 360 & 20& \textbf{10} &18   & 680 & 20& \textbf{3} \\
		\hline
		4 & 480 & 30 &\textbf{10} &	16  & 1020 & 30 &\textbf{22}\\
		\hline
		3 & 1860 & 60 &\textbf{48} & 14 & 1610 & 60 &\textbf{28}\\
		\hline
		2.8 &3250 & 100 & \textbf{55}& 12 &2880 & 110 & \textbf{51}  \\
		\hline
		2.6 & 6120 & 180 & \textbf{100} &10 & 7950 & 250 & \textbf{128} \\
		\hline
		2.3 & - & 1080 & \textbf{577} & 	9.95 & 17620 & 640 & \textbf{361} \\
		\hline
		2.2 & - &5660 & \textbf{4428}&	9.90 & - & 1550 &  \textbf{984}\\
		\hline
	\end{tabular}
	\caption{MSE performance (training error) versus communication cost on the Twitter dataset (large) and the Tom's hardware dataset. For both datasets, COKE saves around 50\% communication resource than DKLA to achieve the same level of learning performance.} 	
\end{table}

\begin{table}
	\centering
	\label{tab:Energy}	 	
	\begin{tabular}{||c||c|c|c||c|c|c||}
		\hline
		& \multicolumn{3}{|c||}{Training error (MSE ($10^{-3}$)) / Commun. cost} &\multicolumn{3}{|c||}{Test error (MSE($10^{-3}$))}\\
		\hline
		Iteration& CTA  &DKLA   &COKE       & CTA  & DKLA   & COKE         \\
		\hline 	 
		$k=50$   &25.65/500   & 22.52/500   &25.22/\textbf{0} 			   &26.45 &22.97 &26.02    \\
		\hline
		$k=100$  &24.88/1000  & 22.12/1000  &23.65/\textbf{57} 			   &25.57 &22.50 &24.2     \\
		\hline
		$k=200$  &24.17/2000  & 21.81/2000  &22.57/\textbf{254} 		   &24.77 &22.15 &23.02  \\
		\hline
		$k=500$  &23.40/5000  & 21.55/5000  &21.88/\textbf{987}            &23.92 &21.86 &22.22    \\
		\hline
		$k=1000$ &22.84/10000 & 21.48/10000 &21.51/\textbf{5752}           &23.31 &21.79 &21.82    \\
		\hline
		$k=1500$ &22.54/15000 & 21.47/15000 &\textbf{21.47}/\textbf{10752} &22.97 &21.78 &\textbf{21.78}   \\
		\hline
		$k=2000$ &22.35/20000 & 21.47/20000 &\textbf{21.47}/\textbf{15752} &22.75 &21.78 &\textbf{21.78}   \\
		\hline
	\end{tabular}
	\caption{MSE performance on the Energy dataset, $\sigma=0.1$, $L=100$, $\lambda = 10^{-3}$, stepsize $\eta = 0.99$ for CTA, $\rho=10^{-2}$ for DKLA and COKE, censoring thresholds $h(k) = 0.5\times 0.98^k$. DKLA and COKE achieve better MSE performance than CTA while COKE requires the least communication resource.}
\end{table} 

\begin{table}
	\centering
	\label{tab:Air}	 	
	\begin{tabular}{||c||c|c|c||c|c|c||}
		\hline
		& \multicolumn{3}{|c||}{Training error (MSE ($10^{-3}$)) / Commun. cost} &\multicolumn{3}{|c||}{Test error (MSE($10^{-3}$))} \\
		\hline
		Iteration &CTA      &DKLA      &COKE                    &CTA &DKLA& COKE    \\
		\hline  
		$k=50$   &6.4/500   &1.8/500   &3.7/\textbf{72}             &6.7 &2.1 &4.0            \\
		\hline
		$k=100$  &4.5/1000  &1.6/1000  &2.2/\textbf{172}            &4.8 &1.8 &2.5            \\
		\hline
		$k=200$  &3.2/2000  &1.4/2000  &1.7/\textbf{384}            &3.5 &1.7 &2.0            \\
		\hline
		$k=500$  &2.2/5000  &1.3/5000  &\textbf{1.3}/\textbf{2263}  &2.5 &1.6 &\textbf{1.6}  \\
		\hline
		$k=1000$ &1.7/10000 &1.2/10000 &\textbf{1.2}/\textbf{7263}  &2.0 &1.6 &\textbf{1.6}   \\
		\hline 
		$k=1500$ &1.6/15000 &1.2/15000 &\textbf{1.2}/\textbf{12263} &1.8 &1.6 &\textbf{1.6}   \\
		\hline
		$k=2000$ &1.5/20000 &1.2/20000 &\textbf{1.2}/\textbf{17263} &1.8 &1.6 &\textbf{1.6}   \\
		\hline
	\end{tabular}
	\caption{MSE performance on the Air quality dataset, $\sigma=0.1$, $L=200$, $\lambda = 10^{-5}$, stepsize $\eta = 0.99$ for CTA, $\rho=10^{-2}$ for DKLA and COKE, censoring thresholds $h(k) =  0.9\times 0.97^k$. DKLA and COKE achieve better MSE performance than CTA while COKE requires the least communication resource than DKLA.}	
\end{table} 

\begin{table}
	\centering
	\label{tab:Energy_CommCost}	 	
		\begin{tabular}{||c||c|c|c||c||c|c|c||}
		\hline
		\multicolumn{4}{|c||}{Twitter dataset (large)} & \multicolumn{4}{|c||}{Tom's hardware} \\
		\hline 
	  MSE ($10^{-3}$) 	& \multicolumn{3}{|c||}{Commun. cost} &   MSE ($10^{-3}$) 	& \multicolumn{3}{|c||}{Commun. cost} \\
		\hline
		& CTA    & DKLA    & COKE &	 & CTA    & DKLA    & COKE   \\
		\hline 	 
  25   &860   &20  &\textbf{11} & 5.0 &810   &60   &\textbf{49}\\
		\hline
  24   &2290   &70  &\textbf{48}& 3.0 &2290  &180  &\textbf{81}\\
		\hline
 23.5 &4160   &140  &\textbf{76} & 2.0 &6010  &360  &\textbf{211}\\
		\hline
  23   &7690   &250  &\textbf{134} & 1.8 &8160  &490  &\textbf{285}\\
		\hline
 22.5  &14750   &480   &\textbf{258} &1.6 &12300 &750  &\textbf{424}\\
		\hline
  22   &- &1160 &\textbf{652}&1.5 &16190 &1010 &\textbf{586} \\
		\hline
 21.5 &- &4950 &\textbf{4062} & 1.2 &-     &5990 &\textbf{5383 }\\
		\hline
	\end{tabular}
	\caption{MSE performance (training error) versus communication cost on the Energy dataset and the Air quality dataset. For both datasets, COKE saves around 45\%-55\% communication resource than DKLA to achieve the same level of learning performance.}
\end{table}

\section{Concluding Remarks}
\label{sec:con}
This paper studies the decentralized kernel learning problem under privacy concern and communication constraints for multi-agent systems. Leveraging the random feature mapping, we convert the non-parametric kernel learning problem into a parametric one in the RF space and solve it under the consensus optimization framework by the alternating direction method of multipliers. A censoring strategy is applied to conserve communication resources. Through both theoretical analysis and simulations, we establish that the proposed algorithms not only achieve linear convergence rate but also exhibit effective generalization performance. Thanks to the fixed-size parametric learning model, the proposed algorithms circumvent the curse of dimensionality problem and do not involve raw data exchange among agents. Hence, they can be applied in distributed learning that involve big-data and offer some level of data privacy protection. To cope with dynamic environments and enhance the learning performance, future work will be devoted to decentralized online kernel learning and multi-kernel learning.

\acks{We would like to acknowledge support for this project from the National Science Foundation (NSF grants \#1741338 and \#1939553). }

\newpage

\appendix
\section*{Appendix A. Proof of Theorem 1 and Theorem 2}
\label{app:appdxA}
\noindent
{\bf Proof}. 
	As discussed in Section~\ref{subsec:RF_Dectr}, solving the decentralized kernel learning problem in the RF space~\eqref{eq:empirical_risk_dis_theta} is equivalent to solving the problem~\eqref{eq:empirical_risk_theta_N}. From~\eqref{eq:minimizer_RF}, it is evident that the convergence of the optimal functional $f$ in \eqref{eq:empirical_risk_theta_N} hinges on the convergence of the decision variables $\bm{\theta}$ in the RF space. Since in the RF space, the decision variables are data-independent, the convergence proof of DKLA boils down to proving the convergence of a convex optimization problem solved by ADMM. However, the convergence proof of COKE is nontrivial because of the error caused by the outdated information introduced by the communication censoring strategy. Our proof for both theorems consists of two steps. The first step is to show linear convergence of decision parameters $\bm{\theta}$ for DKLA via Theorem 4 and for COKE via Theorem 5 below, which are derived straightforwardly from~\citep{shi2014linear} and~\citep{liu2019communication}, respectively. The second step is to show how the convergence of $\bm{\theta}$ translates to the convergence of the learned functional, which are the same for both algorithms. 
	
Compared to~\citep{shi2014linear} and~\citep{liu2019communication} that deal with general optimization problems for parametric learning, this work focuses on specific decentralized kernel learning problem which is more challenging in both solution development and theoretical analysis. By leveraging the RF mapping technique, we successfully develop the DKLA algorithm and the COKE algorithm. Noticeably, a direct application of ADMM as in~\citep{shi2014linear} on decentralized kernel learning is infeasible without raw data exchanges. Moreover, we analyze the convergence of the nonlinear functional to be learned and the generalization performance of kernel learning in the decentralized setting. The analysis is built on the work of~\citep{shi2014linear} and~\citep{liu2019communication} but goes further, and it is only attainable because of the adoption of the RF mapping.	
	
For both algorithms, the linear convergence of decision variables in the RF space is based on matrix reformulation of \eqref{eq:empirical_risk_dis_theta}. Define $\bm{\Theta}^\ast:= [\bm{\theta}^\ast, \bm{\theta}^\ast,\dots,\bm{\theta}^\ast]^\top\in \mathbb{R}^{N\times L}$ and $\bm{\varTheta}^{\ast}:=[\bm{\vartheta}^\ast, \bm{\vartheta}^\ast,\dots,\bm{\vartheta}^\ast]^\top\in\mathbb{R}^{N\times L}$ be the optimal primal variables, and $\bm{\beta}^\ast$ be the optimal dual variable. Then, for DKLA, Theorem 4 states that $\{\bm{\Theta}^k\}$ ($\bm{\Theta}^k := [\bm{\theta}_1^k, \bm{\theta}_2^k,\dots,\bm{\theta}_N^k]^\top\in \mathbb{R}^{N\times L}$) is R-linear convergent to the optimal $\bm{\Theta}^\ast$. For detailed proof, see~\citep{shi2014linear}.

\noindent
{\bf Theorem 4} {\it
		\textbf{[Linear convergence of decision variables in DKLA]} 	
		For the optimization problem \eqref{eq:empirical_risk_theta_N}, initialize the dual variables as $\bm{\gamma}_i^0=\mathbf{0},\;\forall i$, with Assumptions~\ref{ass:net_connect} - \ref{ass:strong_convex}, then $\{\bm{\Theta}^k\}$ is R-linearly convergent to the optimal $\mathbf{\Theta}^{\ast}$ when $k$ goes to infinity following from 
		\begin{equation}
		\|\bm{\Theta}^k-\bm{\Theta}^\ast\|_F^2 \leq \frac{1}{m_{\hat{R}}} \left[\rho \|\bm{\varTheta}^{k-1}-\bm{\varTheta}^\ast\|_F^2+\frac{1}{\rho} \|\bm{\beta}^{k-1}-\bm{\beta}^\ast\|_F^2\right],
		\end{equation}
		where $\{(\bm{\varTheta}^k,\bm{\beta}^k)\}$ is Q-linearly convergent to its optimal $\{(\bm{\varTheta}^{\ast},\bm{\beta}^\ast)\}$: 
		\begin{equation}
		\rho \|\bm{\varTheta}^{k}-\bm{\varTheta}^\ast\|_F^2+\frac{1}{\rho} \|\bm{\beta}^{k}-\bm{\beta}^\ast\|_F^2 \leq \frac{1}{1+\delta_d}\left[\rho \|\bm{\varTheta}^{k-1}-\bm{\varTheta}^\ast\|_F^2+\frac{1}{\rho} \|\bm{\beta}^{k-1}-\bm{\beta}^\ast\|_F^2\right]
		\end{equation}
		with
		\begin{equation}
		\delta_d = \min\left\{\frac{(\nu-1)\tilde{\sigma}_{\min}^2(\mathbf{S}_{-})}{\nu \tilde{\sigma}_{\max}^2(\mathbf{S}_{+})},\frac{m_{\hat{R}}} {\frac{\rho}{4} \tilde{\sigma}_{\max}^2(\mathbf{S}_{+}) + \frac{\nu}{\rho}M_{\hat{R}}^2 \tilde{\sigma}_{\min}^2(\mathbf{S}_{-})}\right\}, \nonumber
		\end{equation}
		where $\nu>1$ is an arbitrary constant, $\tilde{\sigma}_{\max}(\mathbf{S}_{+})$ is the maximum singular value of the unsigned incidence matrix $\mathbf{S}_{+}$ of the network, and $\tilde{\sigma}_{\min}^2(\mathbf{S}_{-})$ is the minimum non-zero singular value of
		the signed incidence matrix $\mathbf{S}_{-}$ of the network, $m_{\hat{R}}$ and $M_{\hat{R}}$ are the minimum strong convexity constant of the local cost functions and the maximum Lipschitz constant of the local gradients, respectively. The Q-linear convergence rate of $\{(\bm{\varTheta}^k,\bm{\beta}^k)\}$ to $\{(\bm{\varTheta}^{\ast},\bm{\beta}^\ast)\}$ satisfies
			\begin{equation}
			r_c \leq \sqrt{\frac{1}{1+\delta_d}}.
			\end{equation}
	} \hfill\BlackBox

To achieve linear convergence of decision variables in COKE, choosing appropriate censoring functions is crucial. Moreover, the penalty parameter $\rho$ also needs to satisfy certain conditions, see Theorem 5 for details~\citep{liu2019communication}. 
	
\noindent
{\bf Theorem 5} {\it
		\label{theor:linear_converge_theta_coke}
		\textbf{[Linear convergence of decision variables in COKE]} 	
		For the optimization problem \eqref{eq:empirical_risk_theta_N} with strongly convex local cost functions whose gradients are Lipschitz continuous, initialize the dual variables as $\bm{\gamma}_i^0=\mathbf{0},\;\forall i$, set the censoring thresholds to be $h(k) = v\mu^k$, with $v>0$ and $\mu\in (0,1)$, and choose the penalty parameter $\rho$ such that 
		\begin{equation}
		\label{eq：choose_rho_1}
		\begin{split}
		0&<\rho  <\min\left\{\frac{4m_{\hat{R}}}{\eta_1},\frac{(\nu-1)\tilde{\sigma}_{\min}^2(\mathbf{S}_{-})}{\nu \eta_3 \tilde{\sigma}_{\max}^2(\mathbf{S}_{+})},  \left(\frac{\eta_1}{4}+\frac{\eta_2 \tilde{\sigma}_{\max}^2(\mathbf{S}_{+})}{8}\right)^{-1}\left(m_{\hat{R}}-\frac{\eta_3\nu M_{\hat{R}}^2 }{\tilde{\sigma}_{\min}^2(\mathbf{S}_{-})}    \right) \right\},
		\end{split}
		\end{equation}
		where $\eta_1>0$, $\eta_2>0$, $\eta_3>0$ and $\nu>1$ are arbitrary constants, $m_{\hat{R}}$ and $M_{\hat{R}}$ are the minimum strong convexity constant of the local cost functions and the maximum Lipschitz constant of the local gradients, respectively. $\tilde{\sigma}_{\max}(\mathbf{S}_{+})$ and $\tilde{\sigma}_{\min}^2(\mathbf{S}_{-})$ are the maximum singular value of the unsigned incidence matrix $\mathbf{S}_{+}$ and the minimum non-zero singular value of the signed incidence matrix $\mathbf{S}_{-}$ of the network, respectively. Then, $\{\bm{\Theta}^k\}$ is R-linearly convergent to the optimal $\mathbf{\Theta}^{\ast}$ when $k$ goes to infinity following from.} \hfill\BlackBox

\noindent
\textbf{Remark 3.} For the kernel ridge regression problem~\eqref{eq:obj_RF}, the minimum strong convexity constant of the local cost functions and the maximum Lipschitz constant of the local gradients are $m_{\hat{R}}:=\min_i \tilde{\sigma}_{\min}^2(\frac{1}{T_i}\bm{\Phi}_{L}^i(\bm{\Phi}_{L}^i)^\top + \frac{2\lambda}{N} \mathbf{I})$ and $M_{\hat{R}}:=\max_i \tilde{\sigma}_{\max}^2(\frac{1}{T_i}\bm{\Phi}_{L}^i(\bm{\Phi}_{L}^i)^\top + \frac{2\lambda}{N} \mathbf{I})$, respectively.

	With the convergence of decision variables in the RF space given in Theorem 4 and Theorem 5, the second step is to prove the linear convergence of the learned functional $\hat{f}_{\bm{\theta}_i^k}(\mathbf{x})$ to the optimal $\hat{f}_{\bm{\theta}^\ast} (\mathbf{x})$, which is straightforward for both algorithms. 

	Denote $\hat{\bm{f}}_{\bm{\Theta}^k}(\mathbf{x}) = [\hat{f}_{\bm{\theta}_1^k}(\mathbf{x}),\dots, \hat{f}_{\bm{\theta}_N^k}(\mathbf{x})]^\top = \bm{\Theta}^k \bm{\phi}_L(\mathbf{x}) $ and $\hat{\bm{f}}_{\bm{\Theta}^\ast}(\mathbf{x}) = [\hat{f}_{\bm{\theta}^\ast}(\mathbf{x}),\dots, \hat{f}_{\bm{\theta}^\ast}(\mathbf{x})]^\top = \bm{\Theta}^\ast \bm{\phi}_L(\mathbf{x})$, then we have
 
	\begin{equation}
	\begin{split}
	\|\hat{\bm{f}}_{\bm{\Theta}^k}(\mathbf{x}) - \hat{\bm{f}}_{\bm{\Theta}^\ast}(\mathbf{x}) \|_2   &= \|\bm{\Theta}^k \bm{\phi}_L(\mathbf{x})  - \bm{\Theta}^\ast \bm{\phi}_L(\mathbf{x}) \|_2  \\    
	&\leq  \|\bm{\Theta}^k - \bm{\Theta}^\ast  \|_2  \|\bm{\phi}_L(\mathbf{x})\|_2 \\
	&\leq  \|\bm{\Theta}^k - \bm{\Theta}^\ast\|_2, \\
	\end{split}
	\end{equation} 
	where the second inequality comes from the fact that $\|\bm{\phi}_L(\mathbf{x}) \|_2\leq 1$ with the adopted RF mapping. 
	
	For DKLA, we have 
	\begin{equation}
	\begin{split}
	\|\hat{\bm{f}}_{\bm{\Theta}^k}(\mathbf{x}) - \hat{\bm{f}}_{\bm{\Theta}^\ast}(\mathbf{x}) \|_2   
	& \leq  \|\bm{\Theta}^k - \bm{\Theta}^\ast\|_2   \leq \frac{1}{m_{\hat{R}}} \left[\rho \|\bm{\varTheta}^{k-1}-\bm{\varTheta}^\ast\|_F^2+\frac{1}{\rho} \|\bm{\beta}^{k-1}-\bm{\beta}^\ast\|_F^2\right].
	\end{split}
	\end{equation}
	Therefore, the Q-linear convergence of $\{\bm{\varTheta}^{k},\bm{\beta}^k\}$ to the optimal $(\bm{\varTheta}^\ast,\bm{\beta}^\ast)$ translates to the R-linear convergence of $\{\hat{\bm{f}}_{\bm{\Theta}^k}(\mathbf{x})\}$. Similarly, the R-linear convergence of $\{\bm{\Theta}^k\}$ to the optimal $\bm{\Theta}^\ast$ of COKE can be translated from the Q-linear convergence of $\{\bm{\varTheta}^{k},\bm{\beta}^k\}$ to the optimal $(\bm{\varTheta}^\ast,\bm{\beta}^\ast)$, see \citet{liu2019communication} for detailed proof. 
 
	It is then straightforward to see that the individually learned functionals converge to the optimal one when $k$ goes to infinity, i.e., for $i\in\mathcal{N}$,
	\begin{equation}
	\begin{split}
	\lim_{k\rightarrow\infty} |\hat{f}_{\bm{\theta}_i^k}(\mathbf{x}) - \hat{f}_{\bm{\theta}^\ast}(\mathbf{x})| &=\lim_{k\rightarrow\infty} |(\bm{\theta}_i^k)^\top \bm{\phi}_L(\mathbf{x}) -(\bm{\theta}^\ast)^\top \bm{\phi}_L(\mathbf{x})| \\
	&\leq     \lim_{k\rightarrow\infty} \|\bm{\theta}_i^k- \bm{\theta}^\ast \|_2 \|\bm{\phi}_L(\mathbf{x}) \|_2\\
	&\leq    \lim_{k\rightarrow\infty}  \|\bm{\theta}_i^k- \bm{\theta}^\ast\|_2 \\
	& = 0.
	\end{split}
	\end{equation}

\section*{Appendix B. Proof of Theorem 3}
\label{app:appdxB} 
\noindent
{\bf Proof}. The empirical risk \eqref{eq:empirical_alpha_N} to be minimized for the kernel regression problem in the RKHS is
	\begin{equation}
	\label{eq:obj_alpha}
	\underset{\bm{\alpha}\in \mathbb{R}^T}{\min} \;\hat{R}(\bm{\alpha}) = \sum_{i=1}^N\hat{R}_i(\bm{\alpha})= \sum_{i=1}^N \left(\frac{1}{T_i}\|\mathbf{y}_i-\mathbf{K}_i^\top\bm{\alpha}\|_2^2+\lambda_i \bm{\alpha}^\top\mathbf{K} \bm{\alpha} \right),
	\end{equation}
	where $\mathbf{y}_i = [y_{i,1},\dots,y_{i,T_i}]^\top \in \mathbb{R}^{T_i\times 1}$, the matrices $\mathbf{K}_i\in\mathbb{R}^{T\times T_i}$ and $\mathbf{K}\in \mathbb{R}^{T\times T}$ are used to store the similarity of the total data and data from agent $i$, and the similarity of all data, respectively, with the assumption that all data are available to all agents. The optimal solution is given in closed form by
	\begin{equation}
	\label{eq:optimal_estimator_alpha}
	\bm{\alpha}^\ast = (\tilde{\mathbf{K}}^\top \tilde{\mathbf{K}} +\lambda \mathbf{K})^{-1}\tilde{\mathbf{K}}\tilde{\mathbf{y}},
	\end{equation}
	where $\tilde{\mathbf{K}}=[\tilde{\mathbf{K}}_1,\dots,\tilde{\mathbf{K}}_N]\in \mathbb{R}^{T\times T}$ with $\tilde{\mathbf{K}}_i =\frac{1}{\sqrt{T_i}}\mathbf{K}_i$, $\forall i\in\mathcal{N}$, $\tilde{\mathbf{y}}=[\tilde{\mathbf{y}}_1;\dots; \tilde{\mathbf{y}}_N] \in \mathbb{R}^{T \times 1}$ with $\tilde{\mathbf{y}}_i =\frac{1}{\sqrt{T_i}}\mathbf{y}_i$, $\forall i\in\mathcal{N}$, and $\lambda=\sum_{i=1}^N \lambda_i$. Denote the predicted values on the training examples using $\bm{\alpha}^\ast$ as $\mathbf{f}_{\bm{\alpha}^\ast}^i\in \mathbb{R}^{T_i}$ for node $i$ and the overall predictions as $\mathbf{f}_{\bm{\alpha}^\ast} = [\mathbf{f}_{\bm{\alpha}^\ast}^1;\dots; \mathbf{f}_{\bm{\alpha}^\ast}^N]\in \mathbb{R}^T$. 
	In the corresponding RF space, we can denote the predicted values obtained for node $i$ by $\bm{\theta}^\ast$ in \eqref{eq:optimal_KRR_RF} as $\mathbf{f}_{\bm{\theta}^\ast}^i\in \mathbb{R}^{T_i}$ and the overall prediction by $\mathbf{f}_{\bm{\theta}^\ast} =[\mathbf{f}_{\bm{\theta}^\ast}^1 ;\dots;\mathbf{f}_{\bm{\theta}^\ast}^N]\in \mathbb{R}^T$.

	To prove Theorem 3, we start by customizing several lemmas and theorems from the literature, which facilitate proving our main results. 
	\begin{definition}\citep[Definition 2]{bartlett2002rademacher}
		\label{def:Rademacher} 
		Let $\{\mathbf{x}_q\}_{q=1}^Q$ be i.i.d samples drawn from the probability distribution $p_\mathcal{X}$. Let $\mathcal{F}$ be a class of functions that map $\mathcal{X}$ to $\mathbb{R}$. Define the random variable
		\begin{equation}
		\label{eq:empir_rademacher}
		\hat{\mathfrak{R}}_Q(\mathcal{F}) := \mathbb{E}_\epsilon\left[\underset{f\in\mathcal{F}}{\sup}\left|\frac{2}{Q}\sum_{q=1}^Q\epsilon_q f(\mathbf{x}_q)\right||\mathbf{x}_1,\dots,\mathbf{x}_Q\right],
		\end{equation}
		where $\{\epsilon_q\}_{q=1}^Q$ are i.i.d. $\{\pm 1\}$-valued random variables with $\mathbb{P}(\epsilon_q=1)=\mathbb{P}(\epsilon_q=-1)=\frac{1}{2}$. Then, the \textit{Rademacher complexity} of $\mathcal{F}$ is defined as
		\begin{equation}
		\label{eq:rademacher}
		\mathfrak{R}_Q(\mathcal{F}) := \mathbb{E}\left[\hat{\mathfrak{R}}_Q(\mathcal{F})\right]. 
		\end{equation}
	\end{definition}
	
	Rademacher complexity is adopted in machine learning and theory of computation to measure the richness of a class of real-valued functions with respect to a probability distribution. Here we adopt it to measure the richness of functions defined in the RKHS  induced by the positive definite kernel $\kappa$ with respect to the sample distribution $p$.
	
	\begin{lemma}\citep[Lemma 22]{bartlett2002rademacher}
		\label{lemma:Rad_ker_matrix}
		Let $\mathcal{H}$ be a RKHS associated with a positive definite kernel $\kappa$ that maps $\mathcal{X}$ to $\mathbb{R}$. Then, we have $\hat{\mathfrak{R}}_Q(\mathcal{H}) \leq \frac{2}{Q}\sqrt{Tr(\mathbf{K})}$, where $\mathbf{K}$ is the kernel matrix for kernel $\kappa$ over the i.i.d. sample set $\{\mathbf{x}_q\}_{q=1}^Q$. Correspondingly, the Rademacher complexity satisfies $\mathfrak{R}_Q(\mathcal{H}) \leq \frac{2}{Q}\mathbb{E}\left[\sqrt{Tr(\mathbf{K})}\right]$.  
	\end{lemma}	
	
	The next theorem states that the generalization performance of a particular estimator in $\mathcal{H}$ not only depends on the number of data points, but also depends on the complexity of $\mathcal{H}$.

\noindent
{\bf Theorem 6} {\it \citep[Theorem 8, Theorem 12]{bartlett2002rademacher}
		\label{theor:Rademacher_complexity}
		Let $\{\mathbf{x}_q,y_q\}_{q=1}^Q$ be i.i.d samples drawn from the distribution $p$ defined on $\mathcal{X}\times\mathcal{Y}$. Assume the loss function $\ell:\mathcal{Y}\times \mathbb{R}\rightarrow [0,1]$ is Lipschitz continuous with a Lipschitz constant $M_{\ell}$. Define the expected risk for all $f\in\mathcal{H}$ be $\mathcal{E}(f)=\mathbb{E}_p\left[\ell(f(\mathbf{x}),y) \right]$, and its corresponding empirical risk be $\hat{\mathcal{E}}(f) = \frac{1}{Q}\sum_{q=1}^{Q}\ell(y_q,f(\mathbf{x}_q))$. Then, for $\delta_p\in(0,1)$, with probability at least $1-\delta_p$, every $f\in \mathcal{H}$ satisfies
		\begin{equation}
		\label{eq:loss_general}
		\mathcal{E}(f)\leq \hat{\mathcal{E}}(f)+  \mathfrak{R}_Q(\tilde{\ell} \circ \mathcal{H})   +\sqrt{\frac{8\log(2/\delta_p)}{Q}}.
		\end{equation}
		where $\tilde{\ell} \circ \mathcal{H} = \{(\mathbf{x},y)\rightarrow \ell(y,f(\mathbf{x}))-\ell(y,0) |f\in\mathcal{H}\}$.} \hfill\BlackBox

\noindent
{\bf Theorem 7} {\it\citep[Theorem 12]{bartlett2002rademacher}	
		\label{theor:Rademacher_complexity_2}	
		If $\ell:\mathcal{Y}\times \mathbb{R}\rightarrow [0,1]$ is Lipschitz with constant $M_{\ell}$ and satisfies $\ell(0) = 0$, then $\mathfrak{R}_Q(\tilde{\ell} \circ \mathcal{H}) \leq 2 M_{\ell}\mathfrak{R}_Q(\mathcal{H})$.} \hfill\BlackBox

	\begin{lemma}~\citep[Modified Proposition 1]{li2018towards}
		\label{lemma:RKHS_k}
		For the RKHS induced by the kernel $\kappa$ with expression \eqref{eq:kern_fouri}, define $\hat{\mathcal{H}}^k:=\{\hat{f}^k:\hat{f}^k=(\bm{\theta}^k)^\top \bm{\phi}_L(\mathbf{x})=  \sum_{l=1}^L\theta_l^k \phi(\mathbf{x},\bm{\omega}_l)$, then we have $\forall \hat{f}^k\in\hat{\mathcal{H}}^k, \|\hat{f}^k\|_{\hat{\mathcal{H}}^k}^2\leq  \|\bm{\theta}^k\|_2^2$, where $\hat{\mathcal{H}}^k$ is the RKHS of functions $\hat{f}^k$ at the $k$-th step. The kernel that induces $\hat{\mathcal{H}}^k$ is the approximated kernel $\hat{\kappa}_L$ defined in \eqref{eq:kernel_map_L}. 
	\end{lemma}

	\begin{lemma}~\citep[Lemma 6]{li2018towards}
		\label{lem:lemma6}
		For the decentralized kernel regression problem defined in Section \ref{sec:prosta}, let $\mathbf{f}_{\bm{\alpha}^\ast}$, $\mathbf{f}_{\bm{\theta}^\ast}$ be the predictions obtained by~\eqref{eq:optimal_estimator_alpha} and \eqref{eq:optimal_KRR_RF}, respectively. Then, we have 
		\begin{equation}
		\langle \mathbf{y}-\mathbf{f}_{\bm{\alpha}^\ast}, \mathbf{f}_{\bm{\theta}^\ast}-\mathbf{f}_{\bm{\alpha}^\ast} \rangle = 0.
		\end{equation}
	\end{lemma}

\noindent
{\bf Theorem 8}{\it ~\citep[Modified Theorem 5]{li2018towards}
\label{theor:f_x}
For the decentralized kernel regression problem defined in Section \ref{sec:prosta}, let $\lambda_\mathbf{K}$ be the largest eigenvalue of the kernel matrix $\mathbf{K}$, and choose the regularization parameter $\lambda<\lambda_\mathbf{K}/T$ so as to control overfitting. Then, for all $\delta_p \in (0,1)$ and $\|f\|_\mathcal{H}\leq 1$, if the number of random features $L$ satisfies 
\begin{equation}
\label{eq:RF_number_KRR_spectr}
L \geq  \frac{1}{\lambda}(\frac{1}{\epsilon^2}+\frac{2}{3\epsilon})\log \frac{16d_{\mathbf{K}}^\lambda}{\delta_p}\nonumber,
\end{equation} 
then with probability at least $1-\delta_p$, the following equation holds
\begin{equation}
\underset{\|f\|_\mathcal{H}\leq 1}{\sup} \;\underset{\|\bm{\theta}\|\leq \sqrt{2/L}}{\inf} \frac{1}{T}\|\mathbf{f}_\mathbf{x}-\mathbf{f}_{\bm{\theta}^\ast}\|_2^2\leq 2\lambda, 
\end{equation}
where $\mathbf{f}_\mathbf{x}\in\mathbb{R}^T$ is the predictions evaluated by $f_\mathcal{H}$ on all samples and $\epsilon\in(0,1)$.} \hfill\BlackBox
	
With the above lemmas and theorems, we are ready to prove Theorem 3, which relies on the following decomposition:   
\begin{equation}
\label{eq:decomp}
\begin{split}
\mathcal{E}(\hat{f}^k) - \mathcal{E}(f_{\mathcal{H}}) 
& = \underbrace{ \mathcal{E}(\hat{f}^k) - \hat{\mathcal{E}}(\hat{f}^k)}_{\text{(1)\; estimation error}} 
\quad +\quad \underbrace{\hat{\mathcal{E}}(\hat{f}^k) - \hat{\mathcal{E}}(\hat{f}_{\bm{\theta}^\ast})}_{\text{(2) \; convergence error}} \quad+\underbrace{ \hat{\mathcal{E}}(\hat{f}_{\bm{\theta}^\ast}) - \hat{\mathcal{E}}(\hat{f}_{\bm{\alpha}^\ast})}_{\text{(3) \;approximation error of RF mapping}}  \\
&\quad    + \quad \underbrace{ \hat{\mathcal{E}}(\hat{f}_{\bm{\alpha}^\ast}) - \mathcal{E}(\hat{f}_{\bm{\alpha}^\ast})}_{\text{(4)\; estimation error}} 
\quad +  \underbrace{\mathcal{E}(\hat{f}_{\bm{\alpha}^\ast}) -   \mathcal{E}(f_{\mathcal{H}})}_{(5)\;\text{approximation error of kernel representation}},
\end{split}
\end{equation}
where $\mathcal{E}(\hat{f}^k)$, $\hat{\mathcal{E}}(\hat{f}^k)$, 
$\mathcal{E}(\hat{f}_{\bm{\theta}^\ast})$, $\hat{\mathcal{E}}(\hat{f}_{\bm{\theta}^\ast})$, $\hat{\mathcal{E}}(\hat{f}_{\bm{\alpha}^\ast})$, $\mathcal{E}(\hat{f}_{\bm{\alpha}^\ast})$ are defined as follows for the kernel regression problem: 
\begin{align}
\mathcal{E}(\hat{f}^k)&:= \sum_{i=1}^N \mathcal{E}_i(\hat{f}_{\bm{\theta}_i^k}) = \sum_{i=1}^N \mathbb{E}_p [(y-(\bm{\theta}_i^k)^\top \bm{\phi}_L(\mathbf{x}))^2]:= \mathbb{E}_p [\|\mathbf{y}_N - \bm{\Phi}_N\tilde{\bm{\Theta}}^k\|_2^2], \nonumber\\
\hat{\mathcal{E}}(\hat{f}^k) &:= \sum_{i=1}^N \hat{\mathcal{E}}_i(\hat{f}_{\bm{\theta}_i^k}) =\sum_{i=1}^N  \frac{1}{T_i} \sum_{t=1}^{T_i} \|\mathbf{y}_i-(\bm{\Phi}_{L}^i)^\top\bm{\theta}_i^k\|_2^2 = \sum_{i=1}^N  \|\tilde{\mathbf{y}}_i-(\tilde{\bm{\Phi}}_{L}^i)^\top\bm{\theta}_i^k\|_2^2=\|\tilde{\mathbf{y}}- \tilde{\bm{\Phi}}_B\tilde{\bm{\Theta}}^k\|_2^2, \nonumber \\
\hat{\mathcal{E}}(\hat{f}_{\bm{\theta}^\ast}) &:= \sum_{i=1}^N \hat{\mathcal{E}}_i(\hat{f}_{\bm{\theta}^\ast}) =\sum_{i=1}^N  \frac{1}{T_i} \sum_{t=1}^{T_i} \|\mathbf{y}_i-(\bm{\Phi}_{L}^i)^\top\bm{\theta}^\ast\|_2^2 = \sum_{i=1}^N  \|\tilde{\mathbf{y}}_i-(\tilde{\bm{\Phi}}_{L}^i)^\top\bm{\theta}^\ast\|_2^2=\|\tilde{\mathbf{y}}- \tilde{\bm{\Phi}}_B\tilde{\bm{\Theta}}^\ast\|_2^2, \nonumber \\ 
\mathcal{E}(\hat{f}_{\bm{\alpha}^\ast}) & := \sum_{i=1}^N \hat{\mathcal{E}}_i(\hat{f}_{\bm{\alpha}^\ast}) = \sum_{i=1}^N \mathbb{E}_p[(y-(\bm{\alpha}^\ast)^\top\bm{\kappa}(\mathbf{x}))^2 ]:= \mathbb{E}_p[(y-(\bm{\alpha}^\ast)^\top\bm{\kappa}(\mathbf{x}))^2 ],  \nonumber\\
\hat{\mathcal{E}}(\hat{f}_{\bm{\alpha}^\ast}) &:= \sum_{i=1}^N  \frac{1}{T_i} \|\mathbf{y}_i-\mathbf{K}_i\bm{\alpha}^\ast\|_2^2   = \sum_{i=1}^N  \|\tilde{\mathbf{y}}_i-\tilde{\mathbf{K}}_i\bm{\alpha}^\ast\|_2^2, \nonumber 
\end{align}
where $\mathbf{y}_N = y\mathbf{1}_N$, 
$\bm{\Phi}_N=
\begin{bmatrix}
\bm{\phi}_L(\mathbf{x}) &\cdots& \mathbf{0}\\
\vdots &\ddots  &\vdots\\
\mathbf{0}&\cdots& \bm{\phi}_L(\mathbf{x})
\end{bmatrix}\in\mathbb{R}^{N\times NL}$, 
$\tilde{\bm{\Phi}}_B=
\begin{bmatrix}
(\tilde{\bm{\Phi}}_L^1)^\top&\cdots& \mathbf{0}\\
\vdots &\ddots  &\vdots\\
\mathbf{0}&\cdots&(\tilde{\bm{\Phi}}_L^N)^\top
\end{bmatrix}\in\mathbb{R}^{T\times NL}$, 
$\tilde{\bm{\Theta}}^k =[\bm{\theta}_1^k;\dots;\bm{\theta}_i^k]\in\mathbb{R}^{NL}$, 
and $\tilde{\bm{\Theta}}^\ast =[\bm{\theta}^\ast;\dots;\bm{\theta}^\ast]\in\mathbb{R}^{NL}$. 
	
Then, we upper bound the excessive risk of $\mathcal{E}(\hat{f}^k)$ learned by COKE by upper bounding the decomposed five terms. For term (1), for $\delta_{p_1} \in(0,1)$, with probability at least $1-\delta_{p_1}$, we have
\begin{equation}
\label{eq:risk_1}
\begin{split}
\mathcal{E}(\hat{f}^k)- \hat{\mathcal{E}}(\hat{f}^k)
&\leq 2M_{\ell_1} \mathfrak{R}_{T}(\tilde{\ell}_1 \circ \hat{\mathcal{H}}^k) + \sqrt{\frac{8\log(2/\delta_{p_1})}{T}} \\
& \leq    2 M_{\ell_1}\mathfrak{R}_{T}(\hat{\mathcal{H}}^k) + \sqrt{\frac{8\log(2/\delta_{p_1})}{T}} \\
&\leq     \frac{4 M_{\ell_1}}{T} \mathbb{E}[Tr(\hat{\mathbf{K}})] + \sqrt{\frac{8\log(2/\delta_{p_1})}{T}} \\
& \leq    \frac{4 M_{\ell_1}}{T} \sqrt{T} + \sqrt{\frac{8\log(2/\delta_{p_1})}{T}} \\
&=  \frac{C_1}{\sqrt{T}},
\end{split}
\end{equation} 
where $C_1:= 4M_{\ell_1}+\sqrt{8\log(2/\delta_{p_1})}$, and $M_{\ell_1}$ is the Lipschitz constant for loss function $\ell_1(\hat{f}_{\bm{\theta}_i^k},y)=((\bm{\theta}_i^k)^\top\bm{\phi}_L(\mathbf{x})-y)^2$. The first inequality comes from Theorem 6, the second inequality comes from Theorem 7, and the third inequality comes from Lemma \ref{lemma:Rad_ker_matrix}. For the last inequality, each element in the Gram matrix $\hat{\mathbf{K}}\in\mathbb{R}^{T\times T}$ is given by \eqref{eq:kernel_map_L}, thus $Tr(\hat{\mathbf{K}})\leq T \|\bm{\phi}_L(\mathbf{x})\|_2^2 \leq T$ with the adopted RF mapping such that $\|\bm{\phi}_L(\mathbf{x})\|_2^2\leq 1$.   

Similarly, for term (4), with probability at least $1-\delta_{p_2}$ for $\delta_{p_2} \in(0,1)$, the following holds,
\begin{equation}
\label{eq:risk_4}
\begin{split}
\hat{\mathcal{E}}(\hat{f}_{\bm{\alpha}^\ast}) - \mathcal{E}(\hat{f}_{\bm{\alpha}^\ast})
& \leq   2M_{\ell_2} \mathfrak{R}_{T}(\tilde{\ell}_2 \circ \mathcal{H}) + \sqrt{\frac{8\log(2/\delta_{p_2})}{T}} \\
& \leq  2 M_{\ell_2}\mathfrak{R}_{T}(\mathcal{H}) + \sqrt{\frac{8\log(2/\delta_{p_2})}{T}} \\
&\leq   \frac{4 M_{\ell_2}}{T} \mathbb{E}[Tr(\mathbf{K})] + \sqrt{\frac{8\log(2/\delta_{p_2})}{T}} \\
& \leq    \frac{4 M_{\ell_2}}{T} \sqrt{T} + \sqrt{\frac{8\log(2/\delta_{p_2})}{T}}  \\
& =    \frac{C_2}{\sqrt{T}},
\end{split}
\end{equation} 
where $C_2 := 4M_{\ell_2}+\sqrt{8\log(2/\delta_{p_2})}$, and $M_{\ell_2}$ is the Lipschitz constant for the loss function $\ell_2(\hat{f}_{\bm{\alpha}^\ast},y)=((\bm{\alpha}^\ast)^\top\bm{\kappa}(\mathbf{x}) -y)^2$.

For term (2), we have
\begin{equation}
\label{eq:risk_2}
\begin{split}
\hat{\mathcal{E}}(\hat{f}^k) -  \hat{\mathcal{E}}(\hat{f}_{\bm{\theta}^\ast}) 
&= \|\tilde{\mathbf{y}}- \tilde{\bm{\Phi}}_B\tilde{\bm{\Theta}}^k\|_2^2 - \|\tilde{\mathbf{y}}-\tilde{\bm{\Phi}}_B\tilde{\bm{\Theta}}^\ast\|_2^2  \\
&\leq  \nabla \left(\|\tilde{\mathbf{y}}-\tilde{\bm{\Phi}}_B\tilde{\bm{\Theta}}^\ast\|_2^2\right)  \|\tilde{\bm{\Theta}}^k-\tilde{\bm{\Theta}}^\ast\|_2 +\frac{M_{\ell_3}}{2}\|\tilde{\bm{\Theta}}^k-\tilde{\bm{\Theta}}^\ast\|_2\\
& \leq \left(\|\tilde{\bm{\Phi}}_B^\top(\tilde{\bm{\Phi}}_B\tilde{\bm{\Theta}}^\ast - \tilde{\mathbf{y}} ) \|_2  +  \frac{M_{\ell_3}}{2}\right)     \|\tilde{\bm{\Theta}}^k-\tilde{\bm{\Theta}}^\ast\|_2\\
& = C_3 \|\tilde{\bm{\Theta}}^k-\tilde{\bm{\Theta}}^\ast\|_2,
\end{split}
\end{equation} 
where $C_3 :=\|\tilde{\bm{\Phi}}_B^\top(\tilde{\bm{\Phi}}_B\tilde{\bm{\Theta}}^\ast - \tilde{\mathbf{y}} ) \|_2  +  \frac{M_{\ell_3}}{2}$, and $ M_{\ell_3}$ is the Lipschitz constant of the loss function $\ell_3(\tilde{\mathbf{y}},\tilde{\bm{\Theta}}) =\|\tilde{\mathbf{y}}-\tilde{\bm{\Phi}}_B\tilde{\bm{\Theta}}\|_2^2$. From Theorem 4 and 5, we conclude $\{\tilde{\bm{\Theta}}^k\}$ converges linearly to $\tilde{\bm{\Theta}}^\ast$.

Term (3) is the approximation error caused by RF mapping, which is bounded by
\begin{equation}
\label{eq:risk_3}
\begin{split}
\hat{\mathcal{E}}(\hat{f}_{\bm{\theta}^\ast})- \hat{\mathcal{E}}(\hat{f}_{\bm{\alpha}^\ast}) 
&=  \sum_{i=1}^{N}  \|\tilde{\mathbf{y}}_i-(\tilde{\bm{\Phi}}_{L}^i)^\top\bm{\theta}^\ast\|_2^2 - \sum_{i=1}^{N} \|\tilde{\mathbf{y}}_i-\tilde{\mathbf{K}}_i\bm{\alpha}^\ast\|_2^2 \\
&= \|\tilde{\mathbf{y}}-\tilde{\mathbf{f}}_{\bm{\theta}^\ast}\|_2^2 - \|\tilde{\mathbf{y}}-\tilde{\mathbf{f}}_{\bm{\alpha}^\ast}\|_2^2 \\
& =\| (\tilde{\mathbf{y}}-\tilde{\mathbf{f}}_{\bm{\alpha}^\ast}) +(\tilde{\mathbf{f}}_{\bm{\alpha}^\ast}-\tilde{\mathbf{f}}_{\bm{\theta}^\ast})\|_2^2 - \|\tilde{\mathbf{y}}-\tilde{\mathbf{f}}_{\bm{\alpha}^\ast}\|_2^2  \\
& =  \underset{\|\tilde{\mathbf{f}}_{\bm{\theta}}\|}{\inf} \left (\|\tilde{\mathbf{y}}-\tilde{\mathbf{f}}_{\bm{\alpha}^\ast}\|_2^2 + \|\tilde{\mathbf{f}}_{\bm{\alpha}^\ast}-\tilde{\mathbf{f}}_{\bm{\theta}} \|_2^2 + 2\langle \tilde{\mathbf{y}}-\tilde{\mathbf{f}}_{\bm{\alpha}^\ast}, \tilde{\mathbf{f}}_{\bm{\alpha}^\ast}-\tilde{\mathbf{f}}_{\bm{\theta}} \rangle \right) -  \|\tilde{\mathbf{y}}-\tilde{\mathbf{f}}_{\bm{\alpha}^\ast}\|_2^2 \\	
&=   \underset{\|\tilde{\mathbf{f}}_{\bm{\theta}}\|}{\inf}  \|\tilde{\mathbf{f}}_{\bm{\alpha}^\ast}-\tilde{\mathbf{f}}_{\bm{\theta}} \|_2^2 + 2 \underset{\|\tilde{\mathbf{f}}_{\bm{\theta}}\|}{\inf} \langle \tilde{\mathbf{y}}-\tilde{\mathbf{f}}_{\bm{\alpha}^\ast}, \tilde{\mathbf{f}}_{\bm{\alpha}^\ast}-\tilde{\mathbf{f}}_{\bm{\theta}} \rangle \\
&\leq \underset{\|\tilde{\mathbf{f}}_{\bm{\theta}}\|}{\inf}  \|\tilde{\mathbf{f}}_{\bm{\alpha}^\ast}-\tilde{\mathbf{f}}_{\bm{\theta}} \|_2^2 +   2\langle \tilde{\mathbf{y}}-\tilde{\mathbf{f}}_{\bm{\alpha}^\ast},\tilde{\mathbf{f}}_{\bm{\alpha}^\ast}- \tilde{\mathbf{f}}_{\bm{\theta}^\ast} \rangle      \\
& = \underset{\|\tilde{\mathbf{f}}_{\bm{\theta}}\|}{\inf}\|\tilde{\mathbf{f}}_{\bm{\alpha}^\ast} - \tilde{\mathbf{f}}_{\bm{\theta}}\|_2^2 \\
&\leq  \underset{\|\tilde{\mathbf{f}}_\mathbf{x}\|}{\sup} \underset{\|\tilde{\mathbf{f}}_{\bm{\theta}}\|}{\inf}\|\tilde{\mathbf{f}}_\mathbf{x} - \tilde{\mathbf{f}}_{\bm{\theta}} \|_2^2\\
&\leq 2\lambda,\\
\end{split}
\end{equation}
where the seventh equality comes from Lemma \ref{lem:lemma6} while the last inequality comes from Theorem 8 with $\tilde{\mathbf{f}}_\mathbf{x}:=[\frac{1}{\sqrt{T_1}}\mathbf{f}_1;\dots;\frac{1}{\sqrt{T_N}}\mathbf{f}_N]\in\mathbb{R}^T$ and $\mathbf{f}_i=[f(\mathbf{x}_{i,1}),\dots, f(\mathbf{x}_{i,T_i})]^\top\in\mathbb{R}^{T_i}$ for $f\in\mathcal{H}$.

To bound term (5) of the approximation error of the models in the RKHS $\mathcal{H}$, we refer to the following Lemma.  

\begin{lemma}\citep[Modified Lemma 5]{rudi2017generalization}
	\label{lem:Rudi_lemma5}
	For the kernel $\kappa$ that can be represented as \eqref{eq:kern_fouri} and bounded RF mapping, that is $\|\phi(\mathbf{x},\bm{\omega})\|\leq 1$ for any $\mathbf{x}\in\mathcal{X}$, under Assumption \ref{ass:f_H-exist}, the following holds for any regularization parameter $\lambda>0$,
	\begin{equation} 
	\mathcal{E}(\hat{f}_{\bm{\alpha}^\ast})-\mathcal{E}(f_{\mathcal{H}}) =	\| \hat{f}_{\bm{\alpha}^\ast} - Pf_p \|_{p_\mathcal{X}}^2\leq (R\lambda^{r})^2\nonumber. 
	\end{equation}
\end{lemma}

In Lemma \ref{lem:Rudi_lemma5}, $f_p$ is the ideal minimizer given the prior knowledge of the marginal distribution $p_\mathcal{X}$ of $\mathbf{x}$ and $P$ is a projection operator on $f_p$ so that $Pf_p$ is the optimal minimizer in RKHS. The parameter $r\in [1/2,1)$ is equivalent to assuming $f_{\mathcal{H}}$ exits, and $R$ can take value as either 1 or $\|\hat{f}_{\bm{\alpha}^\ast}\|_{p_\mathcal{X}}$. Setting $r=1/2$ and $R=1$, we have 
\begin{equation}
\label{eq:risk_5}
\begin{split}
\mathcal{E}(\hat{f}_{\bm{\alpha}^\ast})-\mathcal{E}(f_{\mathcal{H}}) \leq \lambda.
\end{split}
\end{equation}

Combining \eqref{eq:risk_1}-\eqref{eq:risk_5} gives
\begin{equation}
\begin{split}
\lim_{k\rightarrow \infty} \mathcal{E}(\hat{f}^k)-\mathcal{E}(f_{\mathcal{H}}))  
& \leq \lim_{k\rightarrow \infty} \left[\frac{C_1 }{\sqrt{T}} +  C_3\|\tilde{\bm{\Theta}}^k-\tilde{\bm{\Theta}}^\ast\|_2 +2\lambda +  \frac{C_2 }{\sqrt{T}}+\lambda\right]\\
& = \lim_{k\rightarrow \infty}\left[3\lambda+  \frac{C_1+C_2  }{\sqrt{T}}+ C_3\|\tilde{\bm{\Theta}}^k-\tilde{\bm{\Theta}}^\ast\|_2\right]\\
& =  3\lambda + O(\frac{1}{\sqrt{T}}) ,
\end{split}
\end{equation}
and completes the proof.  \hfill\BlackBox


\bibliographystyle{apacite}
\bibliography{references}

\end{document}